\documentclass[hidelinks,onefignum,onetabnum]{siamart220329}

\usepackage{lipsum}
\usepackage{amsfonts}
\usepackage{graphicx}
\usepackage{epstopdf}
\usepackage{algorithmic}
\ifpdf
  \DeclareGraphicsExtensions{.eps,.pdf,.png,.jpg}
\else
  \DeclareGraphicsExtensions{.eps}
\fi

\newsiamremark{remark}{Remark}
\newsiamremark{hypothesis}{Hypothesis}
\crefname{hypothesis}{Hypothesis}{Hypotheses}
\newsiamthm{claim}{Claim}

\headers{PREDICTION COMBINING OFFLINE AND ONLINE LEARNING}{Haizheng Li, Lei Guo}

\title{adaptive prediction theory combining offline and online learning \thanks{This version: Nov 22, 2025.
\funding{This work was supported by the National Natural Science Foundation of China under Grant No. 12288201.}}}

\author{Haizheng Li$^\dagger$ and Lei Guo\thanks{
State Key Laboratory of Mathematical Sciences, Academy of Mathematics and Systems Science, Chinese Academy of Sciences, Beijing 100190, China; School of Mathematical Science, University of Chinese Academy of Sciences, Beijing 100049, China.
(\email{lihaizheng@amss.ac.cn}, \email{Lguo@amss.ac.cn})
}}

\usepackage{amsopn}

\usepackage{array} 
\usepackage[caption=false]{subfig} 
\newcolumntype{R}{>{$}r<{$}} %
\newcolumntype{V}[1]{>{[\;}*{#1}{R@{\;\;}}R<{\;]}} %
\usepackage{amssymb}
 \usepackage{graphicx,epstopdf} 
 \usepackage[caption=false]{subfig} 
\newtheorem{assumption}{Assumption}[section]

\ifpdf
\hypersetup{
  pdftitle={Theory of prediction},
  pdfauthor={Haizheng Li, Lei Guo}
}
\fi

\begin{document}

\maketitle

\begin{abstract}
Real-world intelligence systems usually operate by combining offline learning and online adaptation with highly correlated and non-stationary system data or signals, which, however, has rarely been investigated theoretically in the literature. This paper initiates a theoretical investigation on the prediction performance of a two-stage learning framework combining offline and online algorithms for a class of nonlinear stochastic dynamical systems. For the offline-learning phase, we establish an upper bound on the generalization error for approximate nonlinear-least-squares estimation under general datasets with strong correlation and distribution shift, leveraging the Kullback-Leibler divergence to quantify the distributional discrepancies. For the online-adaptation phase, we address, on the basis of the offline-trained model, the possible uncertain parameter drift in real-world target systems by proposing a meta-LMS prediction algorithm. This two-stage framework, integrating offline learning with online adaptation, demonstrates superior prediction performances compared with either purely offline or online methods. Both theoretical guarantees and empirical studies are provided.
\end{abstract}


{\small 
\section{Introduction}
\subsection{Background} 
The development of big data and large models, particularly deep neural network-based learning methods, have provided many new approaches for the prediction and control of stochastic dynamical systems with uncertainties. Among these, the integration of offline learning and online adaptation is the most widely applied(e.g., \cite{Nature,neuralfly2022neural,MPCyingyong,multimodel2024online}). Offline learning leverages pre-collected historical datasets for model training through batch processing. By consolidating all computationally intensive tasks into a single upfront training phase, this strategy enables efficient and rapid predictions while preserving the capacity to capture essential system dynamics(e.g., \cite{MPCyingyong,neuralfly2022neural}). Furthermore, offline learning from multiple historical trajectories or from multiple environments, unlike single trajectory approaches, enables the acquisition of generalized behavioral patterns across broader operating regimes(e.g., \cite{duotiaoguijihaochu,neuralfly2022neural}). However, this approach often requires excessive storage space and struggles to adapt to dynamically changing environments or unfamiliar scenarios that often arise during real-world deployment. Achieving guaranteed generalization performance may require the smallness of either the parameter sensitivity function or the parameter estimation error, which is an outcome that depends on strong data or model structure conditions such as persistent excitation and expansiveness condition(e.g., \cite{charis2024rate,similarsystem,ziemann2024learning}). When discrepancies exist between real and training environments, the prediction performance often degrades significantly and not robust.
In contrast, online learning processes data, typically acquired from a single system trajectory, in a streaming fashion, requiring minimal storage and computational resources. Its lightweight, streaming processing enables flexibility and scalability. Fortunately, many convergence results have been established for online learning algorithms, particularly when applied to linear and saturated linear stochastic systems in the area of adaptive systems theory(e.g., \cite{guo1995,baohe,l1baohe}). However, online adaptation methods typically face several limitations. They generally require random initialization, which may induce large transient errors(e.g., \cite{yanxia2002stochastic}). Moreover, the theoretical analysis of pure online adaptation for general nonlinear dynamical systems may lead to estimating all the unknown parameters which may otherwise be partly estimated via offline learning. In other words, real-time model online updates usually necessitate full parameter optimization, demanding substantial computational resources that may be challenging to implement in embedded systems or real-time application scenarios. Extensive empirical evidence indicates that full parameter updates often underperform compared to carefully designed partial update strategies, such as linear probing(e.g., \cite{fullbad}).

Given their complementary strengths and weaknesses above, integrating offline and online approaches presents a promising path towards performance advancement. In the era of big data and large models, the increasing complexity of engineering systems, from industrial process control to autonomous navigation and robotic manipulation, has made such offline-online framework not just appealing, but necessary.
The existing representative methods usually follow a two-stage approach: first, they learn either basis functions for unknown system dynamics(e.g., \cite{neuralfly2022neural}) or control policies via deep reinforcement learning(e.g., \cite{Nature}) in idealized environments. Then, during real-time processing, the methods apply online adaptation techniques such as Kalman-filtering or recursive-least-squares to refine model parameters in real time. This two-stage framework begins with extensive offline training on historical or simulated data to develop high-capacity foundation models, which serve as a solid baseline. It is then  followed by online linear-probing in the target system, combining the strengths of both learning approaches to ensure robust and reliable performance under real-world uncertainties(e.g., \cite{MPCyingyong,Nature,neuralfly2022neural,multimodel2024online}).

Despite compelling empirical evidence across multiple domains, there has been limited theoretical investigation on the integration of offline-learning and online-adaptation, particularly in quantitatively characterizing their complementary effects and the resulting overall prediction performance. The state-of-the-art applications still rely primarily on heuristic combination offline and online components, with limited unifying theory offering end-to-end performance guarantees for the complete learning pipeline. Most existing frameworks fundamentally assume that offline learning can sufficiently capture the unknown system dynamics(e.g., \cite{MPCyingyong,neuralfly2022neural}). The core theoretical challenges appear in two lines of sight. First, differences between offline training environments and online deployment conditions cause distribution shift (e.g., \cite{offlineRLconcen}), while parameter drift between source and target systems further violates key assumptions like identical distribution and stationarity that underpin traditional statistical learning theory (e.g., \cite{charis2024rate, ziemann2024learning,singletrajectory2022}). The offline datasets often originate from previous system executions, thus carrying both temporal dependencies that cannot satisfy classical i.i.d. requirements. Apart from this, real-world systems often undergo parameter drift, where system parameters evolve over time due to factors like aging or environmental changes. When combined with distribution shift, this drift results in a substantial mismatch between offline-trained models and their behavior during processing. Existing theoretical frameworks are ill-equipped to capture such coupled dynamics, making it difficult to provide reliable performance guarantees. Thus, it finally renders the learning methods inapplicable to the safety-critical applications like the autonomous driving, weather systems, medical devices, and industrial control fields where traceable and strong performance is mandated.

In this paper, we will provide formal prediction performance guarantees for the complete two-stage learning pipeline and our theoretical development will address the coupled challenges of distribution shift and parameter drift, while providing practical guidance for implementing two-stage learning systems in safety-critical applications.
\subsection{Contributions}
This paper addresses the theoretical foundation of a two-stage offline and online learning framework by making contributions in three aspects: \textbf{(i)Offline-learning phase.} Based on historical trajectories generated by the source system, we estimate the unknown parameter through approximate nonlinear-least-squares and establish a novel upper bound on the generalization error. This bound explicitly addresses non-i.i.d. data with strong correlations and distribution shift, leveraging Kullback-Leibler(KL) divergence to quantify the discrepancy between the training and new data under mild assumptions. This result not only provides theoretical guarantees for the offline-learning generalization in nonlinear dynamical systems with bounded nonlinear mappings(encompassing a class of deep neural networks) but also establishes a foundation for the subsequent online-adaptation phase. \textbf{(ii)Online-adaptation phase.} On the basis of the offline-learning phase, we further propose a meta-LMS prediction algorithm that integrates a multi-model meta-algorithm with a projected least-mean-squares(LMS) algorithm. It is designed to circumvent the parameter drift in the target system and mitigate the excessive transient error inherent in single-model approaches. We then provide a rigorous theoretical analysis of the prediction performance of this two-stage offline and online learning framework with a special focus on cross-system drift in the target system. Our analysis demonstrates that, under certain typical conditions, the framework can achieve near-optimal or even optimal prediction performance.
\textbf{(iii)Comparison with purely offline/online methods.} Finally, we provide a rigorous comparative analysis demonstrating the advantages of the proposed two-stage learning framework over traditional purely offline or online methods. Specifically, we systematically articulate the necessity of online adaptation and the fundamental role of offline learning through both theoretical analysis and experimental validation. 

The remainder of this paper is organized as follows. Section \ref{sec2} introduces the notations and presents the two-stage algorithm, detailing both the offline-learning and online-adaptation phases. Section \ref{sec3} establishes a Generalization Lemma and states the main theorems on the prediction performance, while also examining advantages of our framework relative to conventional purely offline and purely online methods. Proofs of the main theorems are provided in Section \ref{sec4}. Numerical simulations demonstrating effectiveness of the two-stage algorithm are provided in Section \ref{sec5}. Concluding remarks are given in Section \ref{sec6}.
\section{Problem Formulation}\label{sec2}
\subsection{Notation}
In this paper, the symbol $x\in{\mathbb{X}^p}\subseteq\mathbb{R}^p$ represents a $p$-dimensional column vector in $\mathbb{X}^p$, and ${x^\tau}$, $\|x\|$, $\|x\|_1$ and $\|x\|_\infty$ denote its transpose, Euclidean norm, 1-norm and infinity-norm, respectively. The inner product is $\langle x,y\rangle\triangleq x^\tau y$ for $x,y\in\mathbb{X}^p$. The radius of a compact set $\mathcal{M}$ is defined as $\frac{1}{2}\max_{x,x'\in \mathcal{M}}\|x-x'\|$ and is denoted by $R_{\mathcal{M}}$. The minimum(maximum) eigenvalue of a symmetric matrix $A\in \mathbb{R}^{n\times n}$ is denoted by $\lambda_{\min}(A)(\lambda_{\max}(A))$. The spectral radius $\rho(B)$ of a matrix $B$ is defined as the maximum modulus among all its eigenvalues. For a matrix $\Gamma\in\mathbb{R}^{n\times m}$, $\|\Gamma\|$ and $\|\Gamma\|_F$ denote the operator norm induced by the Euclidean norm and Frobenius norm, respectively. The operator $vec(\Gamma)$ denotes the vectorization of \( \Gamma \), obtained by stacking its columns into a single column vector.
In a probability space $(\Omega, \mathcal{F},P)$, $\Omega$ is the sample space, the collection of events is referred to as the $\sigma$-algebra $\mathcal{F}$ on $\Omega$ and $P$ is a probability measure on $(\Omega, \mathcal{F})$.
For an $\mathcal{F}$-measurable set $\mathcal{A}$, its complement $\mathcal{A}^c$ is defined by $\mathcal{A}^c=\Omega-\mathcal{A}$.
The event $\mathcal{A}$ is said to happen almost surely(a.s.) if $P(\mathcal{A})=1$. Expectation and probability with respect to all the randomness of the underlying probability space are denoted by $\mathbb{E}$ and $P$, respectively. Expectation with respect to a random variable $X$ is denoted as $\mathbb{E}_X$ or $\mathbb{E}_{P_X}$, which is defined as :$\mathbb{E}_Xf(X,Y)=\int f(X,Y)dP_X$, and $P_X$ denotes the marginal distribution of $X$. $N(\mu,\sigma^2)$ denotes the Gaussian distribution with mean $\mu$ and variance $\sigma^2$. Let $\{Z_{t}\}_{t=0}^{T-1}$ be a stochastic process with joint distribution $P_Z$. For each pair $(i,j)$, let $P_{Z_{i:j}}$ denote the joint distribution of $\{Z_t\}_{t=i}^j$ and $\mathcal{Z}_{ij} := \sigma\{Z_i, \ldots, Z_j\}$ the $\sigma$-algebra generated by $\{Z_t\}_{t=i}^j$. $D(P\|Q)$ is the Kullback–Leibler(KL) divergence between two distributions $P$ and $Q$ and is defined as: $D(P\|Q)=\int\log(\frac{dP}{dQ})dP$, which is nonnegative and can be $\infty$.

We need the following definitions in our analysis :
\begin{definition}\label{depm} 
The dependency matrix of $\{Z_t\}_{t=0}^{T-1}$ is the matrix $\Gamma_{\mathrm{dep}}(P_{T}) \triangleq \{\Gamma_{ij}\}_{i,j=0}^{T-1}$ where:  
\[
\Gamma_{ij} = \sqrt{2 \sup_{\substack{A \in \mathcal{Z}_{0:i} \\ B \in \mathcal{Z}_{j:T-1}}} |P_{Z_{j:T-1}}(B|A) - P_{Z_{j:T-1}}(B)|},
\]
for $i < j$, $\Gamma_{ii} = 1$, and $\Gamma_{ij} = 0$, for $i > j$.
\end{definition}
\begin{definition}\label{project}
For a given compact set \( D \subseteq \mathbb{R}^m \), the projection operator \(\Pi_D(\cdot)\) is defined as  
\[
\Pi_D(\beta) = \arg\min_{\omega \in D} \|\beta - \omega\|_, \quad \forall \beta \in \mathbb{R}^m.
\]
\end{definition}
\subsection{Offline-learning phase}\label{offlinedescribe}
At this stage, we attempt to estimate the unknown parameter $\alpha^*$ from datasets generated by certain source system through least squares.

Consider the following $N_1$ source system: 
\[y_{t+1,i}=f_t(\alpha^*,\beta_{i}(t), x_{t,i})+v_{t+1,i},\quad i=1,2,\cdots,N_1,\]
where  $\alpha^* \in \mathcal{M}$ is unknown and can be regarded as an ``internal'' parameter,  $\mathcal{M}$ is a known convex and compact set in $\mathbb{R}^n$ with radius $R_{\mathcal{M}}$; the form of $f_t$ is known; the time-varying parameter $\beta_i(t)$ is known and deterministic in the source system, which can be regarded as an ``environmental'' parameter; $x_{t,i} \in \mathbb{X}^p$ is the regression  vector, $y_{t+1,i}\in\mathbb{Y}^q$ is the output and $v_{t+1,i}$ is the noise. Here $\mathbb{X}$ can be a compact set or $\mathbb{R}$. 

In this section, denote $\{x_{0,i},x_{1,i},\cdots,x_{T-1,i}\}\triangleq x_{0:T-1,i}$ and $\{y_{1,i},y_{2,i},\cdots,y_{T,i}\}\triangleq y_{1:T,i}$. Given datasets $\{\hat x_{0:T-1,i},\hat y_{1:T,i}\}_{i=1}^{N_1}$ generated by the trajectories described above, we aim at using least-squares to estimate the true parameters $\alpha^*$ and then derive a generalization error bound with respect to a new dataset $x_{0:T-1}$,  defined as:
\[\frac{1}{N_1T}\sum_{i=1}^{N_1}\sum_{t=0}^{T-1}\mathbb{E}_{x_{0:T-1}}\bigg|\bigg|f_t(\alpha^*,\beta_i(t),x_t)-f_t(\hat\alpha,\beta_i(t),x_t)\bigg|\bigg|^2.\]

Now we list the assumptions underlying our theoretical analysis, each accompanied by some remarks that will be used in the subsequent analysis:
\begin{assumption}\label{ass1}
The $N_1$ different regression vector sets $\hat x_{0:T-1,i}$ and noise sets $v_{1:T,i}$, $i=1,2,\cdots,N_1$, are mutually independent.  
\end{assumption}
\begin{remark}
It should be emphasized that Assumption \ref{ass1} here pertains to datasets of different trajectories, not necessarily requires that the data within any given trajectory to be independent. For example, if $x_{t,i}=(y_{t,i},y_{t-1,i},\cdots,y_{t-p+1,i},u_{t,i},u_{t-1,i},\cdots,u_{t-q+1,i})^\tau$, then Assumption \ref{ass1} holds provided that the initial conditions, input sets $u_{0:T-1,i}$, $i=1,2,\cdots,N_1$, and noise sequences of different system trajectories are independent, but clearly, for any given $i$, $\{x_{t,i},t=0,1,\cdots,T-1\}$ itself is a dependent sequence.
\end{remark}
\begin{assumption}\label{ass3}
There exists $b_1>0,b_1'>0$, $b_2,b_2'\in [0,1)$, such that for any $i=1,2,\cdots,N_1$, \[\|\Gamma_{dep}(P_{T,i})\| ^2 \leq b_1T ^{b_2} ,\quad\|\Gamma_{dep}(P_{T}')\| ^2 \leq b_1'T ^{b_2'},\] 
where $\Gamma_{dep}(\cdot)$ is the dependency matrix of the regression vectors defined in Definition \ref{depm} and $\Gamma_{dep}(P_T')$ denotes the dependency matrix of the new data sequence $x_{0:T-1}$ defined on the same probability space as the training one and $x_{0:T-1}\in \mathbb{X}^{pT}$.  
\end{assumption}
\begin{remark}
This assumption characterizes the degree of dependence among the regression vectors along a single system trajectory either for training data or new data. In cases where the regression vectors are independent, $m$-dependent or $\phi$-mixing with $\sum_{i=1}^\infty\sqrt{\phi(i)}<\infty$, it follows that $b_2=0$; see \cite{charis2024rate,concentrationinequality,ziemann2024learning} for a more detailed discussion.
\end{remark}
\begin{remark}\label{geophi}
We now show that for a general class of nonlinear time series models, one may also have $b_2=0$. Let $x_t=(y_t,y_{t-1},\cdots,y_{t-p+1})^\tau$. Suppose that $|f_t(\alpha^*,\beta_0(t),x)|$ is bounded by a constant $M_f<\infty$ for all $x\in\mathbb{R}^p,t\geq0$, and that the noise $v_{t+1}$ is $\stackrel{\text{i.i.d.}}{\sim} N(0,1)$ and independent of $y_0$, then the regression vector sequence $\{x_t\}$ is a time-inhomogeneous Markov chain. The aforementioned source system can be formalized as:
\begin{align*}
x_{t+1}=F_t(x_t,x_{t-1},\cdots,x_{t-p+1})+V_{t+1}    ,
\end{align*}
where \(V_{t+1}=(v_{t+1},v_{t},\cdots,v_{t-p+2})^\tau\) and $F_t(x_t,x_{t-1},\cdots,x_{t-p+1})=(f_t,f_{t-1},\cdots,f_{t-p+1})^\tau$, $f_s\triangleq f_s(\alpha^*,\beta_0(s),x_s)$, $t-p+1\leq s\leq t$.\\
Since $F_t$ is bounded by $\sqrt{p}M_f$, for any $n\geq 0$, $x_0\in\mathbb{R}^p$, $A\in \mathcal{B}(\mathbb{R}^p)$, we have
\begin{align*}
    P_{n,n+p}(x_0,A)&=P(x_{n+p}\in A|x_n=x_0)\\&\geq \int_{A} (\frac{1}{\sqrt{2\pi}})^p\exp\{\frac{-\|y-F_{n+p-1}\|^2}{2}\}dy_1dy_2\cdots dy_p\\&\geq \delta v(A),
\end{align*}
where $\delta=(\frac{1}{\sqrt{2}})^p\exp\{-pM_f^2\}$ and $v(\cdot)$ is a probability measure induced by $p$-dimensional normal distribution $N(0,\frac{1}{2}I_p)$, thus $\delta v(\cdot)$ is a nonnegative measure with non-zero mass $\delta$.\\
By Remark 1 of Theorem 1 in Chapter 2.4 of \cite{phimixing1994}, this time-inhomogeneous Markov chain $\{x_t\}$ is uniformly geometrically $\phi$-mixing, thus $b_2=0$. 

Moreover, if $x_t=(y_t,y_{t-1},...,y_{t-p+1},u_t,u_{t-1},...,u_{t-q+1})^\tau$, and the inputs $\{u_t\}$ is independent and identically distributed standard normal, a similar argument also shows that $b_2=0$. For a class of closed-loop systems with feedback control that satisfies Assumption \ref{ass3}, see \cite{geocontrol,mpcchen,Meynmarkov}.
\end{remark}
\begin{assumption}\label{ass2}
There exists a constant $L>0$, such that for any $ \alpha',\alpha''\in \mathcal{M},x\in \mathbb{X}^p$ and $t\geq0$, \[\|f_t(\alpha',\beta_i(t),x)-f_t(\alpha'',\beta_i(t),x)\|\leq L\|\alpha'-\alpha''\|.\]
\end{assumption}
\begin{remark}\label{dnn}
This assumption is fairly mild when $\mathbb{X}$ is compact and $\beta_0(t)$ is bounded, and it is also possible to derive a specific value of $L$, which can be interpreted as an upper bound on the sensitivity function of $f_t$ with respect to its unknown parameter $\alpha$, as discussed in \cite{senfunction}. For example, consider a deep neural network $f$ of width $H$ and depth $D$, i.e., $f(\alpha,x)=W_D^\tau\sigma(W_{D-1}^\tau\sigma(\cdots\sigma(W_1^\tau x)\cdots)$, where the activation function $\sigma$ satisfies $\|\sigma(x)-\sigma(y)\|\leq l\|x-y\|$ for any $ x,y\in\mathbb{R}^{H}$(e.g., ReLu, tanh, sigmoid, etc.) and $\alpha=(vec^\tau(W_D),vec^\tau(W_{D-1}),\cdots,vec^\tau(W_1))^\tau$. Denote $h_i\triangleq\sigma(W_{i}^\tau \sigma(\cdots\sigma(W_1^\tau x))\cdots)$. We have
\[\|h_1-h_1'\|\leq l\|W_1-W_1'\|\|x\|,\]
and for $i=2,3,\cdots,D-1$,
\begin{align*}
    \|h_i-h_i'\|&\leq l\|W_i^\tau h_{i-1}-W_i^{'\tau}h_{i-1}'\|\\&\leq l\|h_{i-1}\|\|W_i-W_{i}'\|+l\|W_i'\|\|h_{i-1}-h_{i-1}'\|,
\end{align*}
and,
\begin{align*}
    \|f(\alpha,x)-f(\alpha',x)\|&=\|W_D^\tau h_{D-1}-W_D^{'\tau}h_{D-1}'\|\\&\leq \|W_D-W_{D}'\|\|h_{D-1}\|+l\|W_D'\|\|h_{D-1}-h_{D-1}'\|.
\end{align*}
Besides, by Lipschitz property of $\sigma$, we have $\|\sigma(x)\|\leq l\|x\|+\|\sigma(0)\|$, $\|h_1\|\leq l\|W_1\|\|x\|+\|\sigma(0)\|,\|h_i\|\leq l\|W_i\|\|h_{i-1}\|+\|\sigma(0)\|$. Together with $\|W_i-W_i'\|\leq \|W_i-W_i'\|_F\leq\|\alpha-\alpha'\|$, we know that $f$ satisfies this assumption and the Lipschitz constant $L$ is bounded by a known polynomial of $\|x\|$. Moreover, we emphasize that if in Assumption \ref{ass2}, a time $t$ belongs to $[0,T-1]$ and $L$ depends on $T$, then, the corresponding conclusion remains valid under certain conditions, see Corollary \ref{coro3}.
\end{remark}

Let \(\mathcal{F}_{t,i}\triangleq\sigma\{x_{0,i},x_{1,i},...,x_{t,i},v_{1,i},v_{2,i},\cdots,v_{t,i}\}\) be the $\sigma$-algebra generated by the past regressors and noises up to time $t$ of the $ith$ trajectory.
\begin{assumption}\label{ass4}
 There exists a constant $\sigma_v^2 > 0$, such that for all $\lambda \in \mathbb{R},\|u\|=1,i=1,2,\cdots,N_1$, the following inequality holds:
 \(\mathbb{E}[e^{\lambda \langle v_{t+1,i},u\rangle}|\mathcal{F}_{t,i}] \leq e^{\frac{\lambda^2\sigma_v^2}{2}}\).
\end{assumption}
\begin{remark}\label{noisebound}
It can be concluded that the noise sequence must be a martingale difference sequence by Jensen inequality. Moreover, consider any bounded noise $v_{t+1,i}$ for which the conditional expectation $\mathbb{E}[v_{t+1,i}|\mathcal{F}_{t,i}]=\mu_{t}$ is known. We can then apply a transformation by defining $y'_{t+1}=y_{t+1}-\mu_t, v_{t+1,i}'=v_{t+1,i}-\mu_t$. According to Example 2.4 in Chapter 2 of \cite{Highdimension}, Assumption \ref{ass4} still holds under this transformation. Besides, taking a constant $c>2\sigma_v^2$ and defining $A_{t,i}\triangleq\{v_{t,i}:\|v_{t,i}\|^2\geq c\log t\}$, by Markov inequality, we have $P(A_{t,i}|\mathcal{F}_{t-1,i})\leq \inf_{s>0} \exp\{\frac{s^2\sigma_v^2}{2}-s\sqrt{c\log t}\}=\exp\{-\frac{c\log t}{2\sigma_v^2}\}$, thus $\|v_{t}\|^2=O(\log t)$ by Borel-Cantelli Lemma \cite{shibianxitong}.
\end{remark}
\begin{assumption}\label{ass5}
Given a fixed $\varepsilon^*\geq0$ and $N_1$ datasets $\{\hat x_{0:T-1,i},\hat y_{1:T,i}\}_{i=1}^{N_1}$ generated by these system trajectories, a sub-optimal estimator $\hat\alpha\in \mathcal{M}$ can be obtained in the following sense: 
\begin{align*}
&\quad\frac{1}{N_1T}\sum_{i=1}^{N_1}\sum_{t=0}^{T-1}\bigg|\bigg|\hat y_{t+1,i}-f_t(\hat\alpha,\beta_0(t),\hat x_{t,i})\bigg|\bigg|^2
\\&\leq \min\limits_{\alpha\in \mathcal{M}}\frac{1}{N_1T}\sum_{i=1}^{N_1}\sum_{t=0}^{T-1}\bigg|\bigg|\hat y_{t+1,i}-f_t(\alpha,\beta_0(t),\hat x_{t,i})\bigg|\bigg|^2+\varepsilon^*.  
\end{align*}  
\end{assumption}
\begin{remark}
	Notice that the nonlinear-least-squares estimate over $\mathcal{M}$ satisfies this inequality with $\varepsilon^*=0$. For any given $\varepsilon^*>0$, an estimate satisfying this inequality can be obtained using random search method similar to those proposed in \cite{self1996,randomsearchwang}. Besides, we emphasize that no statistical assumptions on the error $\varepsilon^*$ are required. According to Remark \ref{dnn}, a deep neural network with a suitable class of activation functions satisfies Assumption \ref{ass2}. Therefore, when the form of $f_t$ is unknown, the Universal Approximation Theorem for deep neural networks(e.g., \cite{appdnn}) allows us to reformulate the problem as one of learning the network parameters and $\varepsilon^*$ can be decomposed as $\varepsilon^*=\varepsilon_{app}+\varepsilon_{alg}$, where $\varepsilon_{app}$ denotes the approximation error inherent to network’s capacity, and $\varepsilon_{alg}$ represents the error introduced by certain learning algorithms.
\end{remark}

\subsection{Online-adaptation phase}\label{onlinesec}
Unlike the previously studied source system, some parameters of the actual system may be unknown and subject to drift and the new sequence of regression vectors $x_{0:T-1}$ may exhibit distribution shift compared with those during the offline training. We now give a detailed description.

Let us consider the scalar output case, i.e., $q=1$, for convenience. Consider the following real-world target system:
\[
y_{t+1}=f_t(\alpha^*,\beta(t),x_t)+w_{t+1},
\]
where compared with the source system in the offline-learning phase, the ``environmental'' parameter sequence may differ from that of the source system and thus $\beta(t)\neq\beta_0(t)$, which is now assumed to be unknown; $w_{t+1}$ is either a stochastic or a deterministic bounded perturbation satisfying $|w_{t+1}|\leq W_{\max}<\infty$ for any $t\geq0$, which may not satisfy Assumption \ref{ass4}. In addition, since most real-world systems are subject to fundamental physical laws that inherently guarantee their boundedness, we assume that $|f_t(\alpha^*,\beta(t),x_t)|\leq M_f<\infty$ for any $t\geq0$, even though $f_t(\alpha^*,\beta(t),x_t)$ may not satisfy Assumption \ref{ass2}. 

The constants $M_f$ and $W_{\max}$ are both known a prior, and for simplicity, the new sequence of regression vectors $x_{0:T-1}$ is considered to be independent of the training datasets $\{\hat x_{0:T-1,i},y_{1:T,i}\}_{i=1}^{N_1}$ originating from the source system(see Corollaries \ref{coro2}, \ref{coro3} for the non-independent case).

In this section, the parameter estimate $\hat{\alpha}$ obtained from the offline-learning phase remains fixed, while online adaptation is employed to adapt the unknown, time-varying parameter $\beta(t)$ only.
 Our focus will be put on the prediction performance of the target system output in an average sense:
\[
         J_{T}\triangleq\frac{1}{T}\sum_{t=0}^{T-1}\mathbb{E}(y_{t+1}- y_{t+1}^{\text{pred}})^2,
\]
where $y_{t+1}^{\text{pred}}\in\sigma\{\hat\alpha,x_0,x_1,\cdots,x_t,y_1,y_2,\cdots,y_t\}$ is a predicted value of the unseen $y_{t+1}$, and the expectation is taken over all randomness.

In the following, we consider a typical case where $f_t(\alpha,\beta,x)$ can be linearly parameterized or approximated by a separable model of the form $\sum_{i=1}^m\lambda_i(t)\varphi_i(\beta,t)\psi_i(\alpha,x,t)$, which is hold for quite general classes of nonlinear functions, see, e.g., the related mathematical theorems in \cite{representationsong1,representationsong2,mercer,mercerconvergence}. Thus we may assume that for all $\alpha\in \mathcal{M}$ and $t\geq0$,
\begin{equation}\label{separable}
f_t(\alpha,\beta(t),x_t)=\beta_t^\tau\phi_t(\alpha,x_t),\end{equation}
where $\beta_t$ is an unknown $m$-dimensional vector whose components are the weighting coefficients $\lambda_i(t)\varphi_i(\beta(t),t)$ of the ``basis'' functions $\psi_i(\alpha,x_t,t)$ represented by the components of the vector $\phi_t$. In the presence of approximation error, it can in fact be treated as part of the perturbation, thereby yielding analogous theoretical results.

Therefore, we consider the following real-world target system:
\begin{equation}\label{originalsystem}
    \begin{aligned}
        y_{t+1} = \beta_t^\tau \phi_t(\alpha^*, x_t) + w_{t+1}.
    \end{aligned}
\end{equation}
Additionally, we impose the following assumption:
\begin{assumption}\label{assa1}
There exists known constants $A,B$, such that for the $\hat\alpha\in \mathcal{M}$ obtained from the offline-learning phase, $\|\phi_t(\hat{\alpha}, x_t)\| \leq A$, $\|\beta_t\| \leq B$, for any  $t\geq0$.
\end{assumption}

The original target system (\ref{originalsystem}) can be rewritten as:
\begin{equation}
    \begin{aligned}
        y_{t+1} = \beta_t^\tau \phi_t(\hat{\alpha}, x_t) + \varepsilon_{t} + w_{t+1},
    \end{aligned}
\end{equation}
where $\varepsilon_t\triangleq \beta_t^\tau\phi(\alpha^*,x_t)-\beta_t^\tau\phi(\hat\alpha,x_t)$ represents the error caused by the discrepancy between the parameter $\hat\alpha$ learned through offline training and the true system parameters $\alpha^*$. Under Assumption \ref{assa1}, $\beta_t$ is time-varying and bounded, $\phi_t(\hat{\alpha}, x_t)$ is bounded. 

Similar to (\ref{separable}), we can also write $f_t(\alpha,\beta_0(t),x_t)=\beta_t^{0\tau}\phi_t(\alpha,x_t)$, where $\beta_t^0$ is known when $\beta_0(t)$ is known as assumed in the offline-learning phase.  Furthermore, the error $\varepsilon_t$ is straightforwardly bounded by $M_f+AB$, thus there always exist a constant $L_1\geq0$, and a bounded sequence $L_0(t)\geq0$, such that for all $t\geq0$:
\begin{equation}\label{transfer}
    \begin{aligned}
    \mathbb{E}\varepsilon_{t}^2 \leq L_1 \mathbb{E}\bigg(\beta_t^{0\tau} \phi_t(\alpha^*, x_t) - \beta_t^{0\tau} \phi_t(\hat{\alpha}, x_t)\bigg)^2 + L_0(t).
    \end{aligned}
\end{equation}
This means that the generalization error at time $t$ on the target system is bounded by that on the source system plus an offset term $L_0(t)$.  Moreover, $L_1$ and $L_0(t)$ serve to characterize the parameter drift from the source to the target system, and their specific choices satisfying (\ref{transfer}) are clearly not unique as detailed in Appendix \ref{assbridge}, and will influence the output prediction error bound as discussed in Remark \ref{choices}. As remarked in Case 2 of Appendix \ref{assbridge}, the term $L_0(t)$ can reflect the squared norm of the inner product of the generalization error $\phi_t(\alpha^*,x_t)-\phi_t(\hat\alpha,x_t)$ and the projection residual of $\beta_t$ to the linear space spanned by $\beta_t^0$.

Next, to predict the output at the next step, following previous studies \cite{lmsguo,multimodel2024online,predictionlearning,convexmeta,yanxia2002stochastic}, we propose a meta-LMS prediction Algorithm \ref{pred1}, aiming at enhancing the transient performance through a multi-model framework with prediction-error-based weighting, by employing multiple initializations under a single hyperparameter $d$, a mechanism analogous to the switching strategy in \cite{yanxia2002stochastic}. Unlike the multiple hyperparameters in \cite{convexmeta}, our focus is to generate the predicted outputs rather than the estimated parameters.
\begin{algorithm}
\caption{Meta-LMS Prediction}\label{pred1}
\begin{algorithmic}[1]
\REQUIRE Learning rate $\lambda>0$, $\gamma\in(0,1)$, $d>\frac{A^2}{(1-\gamma)^2}$.
\STATE $w_{0,i}>0$ and $\sum_{i=1}^{N_2}w_{0,i}=1$, $D\triangleq\bigg\{\beta\in\mathbb{R}^m\bigg|\|\beta\|\leq B\bigg\}$, $\|\phi_0(\hat\alpha,x_0)\|\leq m_0\leq \frac{A}{1-\gamma}$, and the initial values $\hat\beta_{0,i}$ of these $N_2$ models are chosen arbitrarily in $D$ and mutually different.
\FOR{$t = 0, 1, \ldots, T-1$}
\STATE Obtain prediction ${y}_{t+1,i}^{\text{pred}} = \hat\beta_{t,i}^\tau\phi_t(\hat\alpha,x_t)$ from each model $i$.
\STATE Aggregate prediction ${y}_{t+1}^{\text{pred}} = \sum_{i=1}^{N_2} w_{t,i} {y}_{t+1,i}^{\text{pred}}.$
\STATE Incur loss $l_t(y_{t+1,i}^{\text{pred}})\triangleq\bigg(y_{t+1}-y_{t+1,i}^{\text{pred}}\bigg)^2$ for each model $i$.
\STATE Update the weights:  
\[
w_{t+1,i} = \frac{w_{t,i} \exp(-\lambda l_t(y_{t+1,i}^{\text{pred}}))}{\sum_{i=1}^{N_2} w_{t,i} \exp(-\lambda l_t(y_{t+1,i}^{\text{pred}}))}
.\]
\STATE Update the estimate of $\beta_{t+1}$ in parallel using projected LMS algorithm:
\[\hat{\beta}_{t+1,i} = \Pi_D \left\{ \hat{\beta}_{t,i}+ \frac{\phi_t(\hat\alpha,x_t)}{d+m_t^2} \left( y_{t+1} - \hat\beta_{t,i}^\tau\phi_t(\hat\alpha,x_t) \right) \right\},\quad i=1,2,\cdots,N_2.\]
\STATE Update $m_{t+1}$:
\[m_{t+1}=\gamma m_{t}+\|\phi_{t+1}(\hat\alpha,x_{t+1})\|.\]
\ENDFOR
\end{algorithmic}
\end{algorithm}

Similarly, we derive the following lemma to highlight the theoretical advantage of multi-models over a single-model baseline: Algorithm \ref{pred1} has a performance floor guaranteed by the worst model and asymptotically achieves the performance of the best model:

\begin{lemma}\label{predlemma}
If the learning rate $\lambda<\frac{1}{2(M_f+AB+W_{\max})^2}$, then the average prediction error produced by meta-LMS prediction Algorithm \ref{pred1} satisfies:
\begin{equation}\label{maxmeta}
    \begin{aligned}
        \frac{1}{T}\sum_{t=0}^{T-1}(y_{t+1}-y_{t+1}^{\text{pred}})^2\leq\max_{1\leq i\leq N_2}\bigg\{\frac{1}{T}\sum_{t=0}^{T-1}(y_{t+1}- y_{t+1,i}^\text{pred})^2\bigg\},\quad a.s.,
    \end{aligned}
\end{equation}
and,
\begin{equation}\label{minmeta}
    \begin{aligned}
        J_T=\frac{1}{T}\sum_{t=0}^{T-1}\mathbb{E}(y_{t+1}-y_{t+1}^{\text{pred}})^2\leq\min\limits_{1\leq i\leq N_2} \bigg\{\frac{1}{T}\sum_{t=0}^{T-1}\mathbb{E}(y_{t+1}- y_{t+1,i}^\text{pred})^2+\frac{\log \frac{1}{w_{0,i}}}{\lambda T}\bigg\}.
    \end{aligned}
\end{equation} 
\end{lemma}

\section{Main Results}\label{sec3}

\subsection{Generalization Lemma}
In Section \ref{offlinedescribe}, an estimate $\hat\alpha$ of the true parameter $\alpha^*$ was obtained by employing an approximate nonlinear-least-squares approach on datasets originating from the source system. We first give the following Generalization Lemma on the bound of the generalization error due to distribution shift between the training data and new data. The proof is supplied in Appendix \ref{gleproof}.

\begin{lemma} \label{gle}
	Under Assumptions \ref{ass1}-\ref{ass5}, the generalization error based on the estimate $\hat\alpha$ obtained in the offline-learning phase can be bounded as follows:
    \begin{equation}\label{G1}
\begin{aligned}
&\qquad\frac{1}{T}\sum_{t=0}^{T-1}\mathbb{E}_{x_{0:T-1}}\bigg|\bigg|f_t(\alpha^*,\beta_0(t),x_t)-f_t(\hat\alpha,\beta_0(t),x_t)\bigg|\bigg|^2\\&= O\bigg(\frac{\log (N_1T)}{N_1T^{1-b
    _2}}+\frac{D(P_{T}\|P_{T}')}{T^{1-b_2'}}\bigg)+32\varepsilon^*,\qquad a.s.,   
\end{aligned}
\end{equation}
and there exists  $N_0>0$ depending on $n,R_{\mathcal{M}}, L, \sigma_v$ and $p$ such that when $N_1 \cdot T>N_0$
\begin{equation}\label{maroffset}
\begin{aligned}
   &\qquad\frac{1}{T}\sum_{t=0}^{T-1}\mathbb{E}\bigg|\bigg|f_t(\alpha^*,\beta_0(t),x_t)-f_t(\hat\alpha,\beta_0(t),x_t)\bigg|\bigg|^2\\&\leq C_1\frac{\log N_1T}{N_1T^{1-b_2}}+8\sigma_v^2C_0\frac{\log N_1T}{N_1T}+8L^2R_{\mathcal{M}}^2b_1\frac{D(P_T\|P_T')}{T^{1-b_2'}}+32\varepsilon^*,
\end{aligned}
\end{equation}
where the constant $C_1$ depends on $n,R_{\mathcal{M}},b_1,L,p$; $C_0$ depends on $n,R_{\mathcal{M}},L,p$. The constants $N_0,C_1$ and $C_0$ can be obtained according to the proof in Appendix \ref{gleproof}. $P_{T}'$ denotes the distribution of the new data $x_{0:T-1}$ and $P_{T}$ denotes the distribution of the training data $\hat x_{0:T-1,i}$. 
\end{lemma}
\begin{remark}
Inspired by \cite{charis2024rate}, this lemma derives an upper bound on the generalization error across multiple source system trajectories, while allowing for distribution shift between the training and new data without assuming Assumptions 4 and 5 from \cite{charis2024rate} which require a kind of persistence of excitation condition on the data and a quadratic identifiability condition. The third term on the right-hand side of (3.1) captures the error due to distribution shift which depends on the data marginal distributions, and the fourth term corresponds to the possible optimization errors in Assumption \ref{ass5}.
\end{remark}
\begin{remark}\label{disandpara}
For many dynamical systems, the distribution shift may arise from possible parameter drift in system dynamics, calling for a more refined analysis of (\ref{G1}) in Generalization Lemma \ref{gle}, see, e.g., Corollaries \ref{coro1}, \ref{coro2} and \ref{coro3} in the subsequent section. As is well known, the KL divergence can sometimes be infinite. To address this issue, alternative measures, such as Total Variation distance, Rényi divergence, and concentration coefficient, have been proposed to quantify the difference between distributions(e.g., \cite{offlineRLconcen, iidgeneralization}). However, these alternatives often require stronger assumptions and may be less effective than the KL divergence under some typical situations, e.g., Corollaries~\ref{coro1}, \ref{coro2} and \ref{coro3}.
\end{remark}

\subsubsection{Corollaries of Generalization Lemma \ref{gle}}\label{glepropo}
We note that, under the i.i.d.\ assumption, i.e., $\hat x_t\stackrel{i.i.d.}{\sim}P_0$ and $x_t\stackrel{i.i.d.}{\sim}P_0'$ for all $t\geq0$, the KL divergence term simplifies to
\(
\frac{D(P_T \| P_T')}{T^{1 - b_2'}} = \frac{D(P_T \| P_T')}{T} = D(P_0 \| P_0'),
\)
and thus may not vanish, as is also noted in Remark~2 of~\cite{iidgeneralization}. Nevertheless, under Assumption \ref{ass2}, the expected generalization error can be further bounded by $L^2\mathbb{E}\|\alpha^*-\hat\alpha\|^2$, which vanishes as $\hat\alpha\to\alpha^*$ in the mean square sense.

Thus, the convergence of the parameter estimators plays an important role in characterizing the robustness of the generalization error. As a trade-off, it may require certain ``excitation'' conditions on the data and/or model structure. We now establish two sufficient conditions for the mean square convergence of the nonlinear-least-squares(NLS) parameter estimation based on Generalization Lemma \ref{gle} with proof given in Appendix \ref{coroproof}.

\begin{corollary}\label{paraest}(``Excitation'' conditions for estimator convergence)
    Suppose that Assumptions \ref{ass2} and \ref{ass4} hold. Then, under either of the following two ``excitation" conditions:  
    \begin{equation}\label{exci1}
        \begin{aligned}
            \quad T^{b_2}\log T&=o\Bigg(\mathbb{E}\min\limits_{\alpha'\in \mathcal{M}}\lambda_{\min}\bigg(\sum_{t=0}^{T-1}F_t(\alpha^*,\alpha')F_t^\tau(\alpha^*,\alpha')\bigg)\Bigg), 
        \end{aligned}
    \end{equation} 
    or,
    \begin{equation}\label{exci2}
        \begin{aligned}
            \quad\log T=o\Bigg(\min\limits_{\alpha'\in \mathcal{M}}\lambda_{\min}\bigg(\sum_{t=0}^{T-1}F_t(\alpha^*,\alpha')F_t^\tau(\alpha^*,\alpha')\bigg)\Bigg)\quad a.s.,
        \end{aligned}
    \end{equation}
    where $F_t(\alpha^*,\alpha')\triangleq\int_0^1\nabla_\alpha f_t(\alpha'+s(\alpha^*-\alpha'),\beta_0(t),\hat x_{t,1})ds$,
    the NLS estimator $\hat\alpha$ satisfies:
    \[\mathbb{E}\|\hat\alpha-\alpha^*\|^2\to0,\quad \text{as } T\to\infty.\]
\end{corollary}
\begin{remark}
    In fact, the almost sure convergence of $\hat\alpha$ to the true parameter $\alpha^*$ together with the corresponding convergence rate can be obtained under Condition (\ref{exci2}), and the proof is supplied in Appendix \ref{coroproof}. For either linear stable model or nonlinear dynamical model with bounded nonlinear mapping, it is known that $b_2=0$(e.g., \cite{charis2024rate,ziemann2024learning} and Remark \ref{geophi}). Conditions (\ref{exci1}) and (\ref{exci2}) are reminiscent of the weakest possible convergence condition for the standard linear-least-squares algorithm, see, \cite{laiwei1982}. 
\end{remark}

We now provide three corollaries that illustrate the application of Generalization Lemma \ref{gle} to three typical classes of nonlinear dynamical systems. For convenience, consider the case of a single trajectory and scalar outputs and let the optimization error $\varepsilon^*=0$, while the general case can be considered similarly. The related proofs are provided in Appendix \ref{coroproof}.
\begin{corollary}\label{coro1}(Generalizability for isomorphic systems)
    Consider the following two systems, and it is the second system that generates the new data:
\[y_{t+1}=f_t(\alpha^*,\beta_0(t),\phi_t,u_t)+w_{t+1},\]
\[y'_{t+1}=f_t(\alpha^*,\beta_0(t),\phi_t',u_t')+w'_{t+1},\]
where
\[\phi_t=(y_{t},y_{t-1},\cdots,y_{t-p+1},u_{t-1},\cdots,u_{t-q+1})^\tau,\]
\[\phi_t'=(y'_{t},y'_{t-1},\cdots,y'_{t-p+1},u'_{t-1},\cdots,u'_{t-q+1})^\tau.\]
Assume that:

(i) $\|f_t(\alpha,\beta_0(t),\phi_,u)\|$ is bounded by $M_f<\infty$ for all $t\geq0$, $\alpha\in \mathcal{M}$ and $(\phi,u)\in \mathbb{R}^{p+q}$; and for any compact set $B\subset \mathbb{R}^{p+q}$, there exists a constant $C_B<\infty$, such that $\|\nabla_{\alpha} f_t(\alpha,\beta_0(t),\phi,u)\|\leq C_B$, for all $t\geq0$, $\alpha\in \mathcal{M}$ and $(\phi,u)\in B$, e.g., a deep neural network in Remark \ref{dnn} with bounded activation functions.

(ii) These two systems are independent of each other (initial values, inputs and noises) and both systems share the same feedback input structure, i.e., $u_{t}=\pi_t(\phi_{t})+v_{t}$, $u_{t}'=\pi_t(\phi_t')+v_{t}'$, and $|\pi_t(\phi)|\leq M_f<\infty$, for all $\phi\in\mathbb{R}^{p+q-1},t\geq0$.

(iii) $\{w_{t+1},v_t\}$ and $\{w_{t+1}',v_t'\}$ are both independent and identically distributed; $w_{t+1}$ and $w_{t+1}'$ are independent of $\sigma\{v_{0:t},\phi_0\}$ and $\sigma\{v'_{0:t},\phi_0'\}$, respectively; $v_t$ and $v_t'$ are independent of $\sigma\{w_{1:t},\phi_0\}$ and $\sigma\{w'_{1:t},\phi_0'\}$, respectively. Besides, they are zero-mean, bounded by $W>2M_f$ and satisfy $\inf\limits_{v\in[-W,W]}p(v)>0$, $\inf\limits_{w\in[-W,W]}p(w)>0$, where $p(\cdot)$ is denoted as the probability density function.

(iv) The KL-divergence of the initial values $x_0\triangleq(\phi_0^\tau,u_0)^\tau,x_0'\triangleq(\phi_0'^\tau,u_0')^\tau$ satisfies $D(D_0\|D_0')<\infty$, where $x_0\sim D_0$ and $x_0'\sim D_0'$.

Let $\hat\alpha$ be the nonlinear-least-squares estimate of $\alpha^*$, i.e.,
\[\hat\alpha\triangleq\arg\min\limits_{\alpha\in \mathcal{M}}\frac{1}{T}\sum_{t=0}^{T-1}\bigg(y_{t+1}-f_t(\alpha,\beta_0(t),\phi_t,u_t)\bigg)^2,\]
where the training dataset $\{\phi_{0:T-1},u_{0:T-1},y_{1:T}\}$ comes from the first system.

Then, the generalization error has the following convergence rate:
\[\frac{1}{T}\sum_{t=0}^{T-1}\mathbb{E}\bigg(f_t(\alpha^*,\beta_0(t),\phi_t',u_t')-f_t(\hat\alpha,\beta_0(t),\phi_t',u_t')\bigg)^2=O\bigg(\frac{\log T}{T}\bigg),\]
where the expectation is taken over all randomness. 
\end{corollary}
\begin{corollary}\label{coro2}(Generalizability for time series models)
Consider the following system with $l$-dimensional outputs:
\[Y_{t+1}=F(\alpha^*,Y_t)+W_{t+1},\]
where the $F:\mathbb{R}^n\times\mathbb{R}^l\to \mathbb{R}^l$ and the noise process $\{W_{t+1}\}$ satisfy assumptions (i) and (iii), respectively, in Corollary \ref{coro1}. Besides, assume that the initial value $Y_0$ is supported on $B\triangleq B_l(0,W-M_f)$, i.e., a ball of radius $W-M_f$ centered at the origin, and has a density $p_0(x)$ satisfying $\int_{\{x|p_0(x)>1\}}p_0(x)\log p_0(x)dx<\infty$.

Let $\hat\alpha$ be the NLS estimate based on the system's outputs $Y_{0:T}$. Then, the generalization error of the next trajectory of length $T$ starting from time $T$ has the following convergence rate:
\[\frac{1}{T}\sum_{t=0}^{T-1}\mathbb{E}_{Y_{T:2T-1}}\bigg\|F(\alpha^*,Y_{t+T})-F(\hat\alpha,Y_{t+T})\bigg\|^2=O\bigg(\frac{\log T}{T}\bigg)\quad a.s.\]
\end{corollary}
\begin{corollary}\label{coro3}(Generalizability under feedback control with learned parameter)
    Consider the following two systems:
\[y_{t+1}=f_t(\beta_0(t),y_t,u_t)+v_{t+1},\quad u_t=\pi_t(\alpha^*,y_t);\]
\[y_{t+1}'=f_t(\beta_0(t),y_t',u_t')+v'_{t+1},\quad u_t'=\pi_t(\hat\alpha,y_t');\]
where $\hat\alpha$ is the NLS estimate of $\alpha^*$ obtained from the first system.

Suppose that:

    (i) $f_t$ is bounded and $\|\nabla_\alpha f_t(\beta_0(t),y,\pi_t(\alpha,y))\|\leq a|y|^b+c$ for some constants $a,b,c$, for all $\alpha\in \mathcal{M},y\in\mathbb{R},t\geq0$, and $b\geq0$ is known(e.g., a deep neural network in Remark \ref{dnn}).
    
    (ii) The noise $v_t$ is independent of $\sigma\{v_{0:t-1},y_0\}$, and $v_t'$ is independent of $\sigma\{v'_{0:t-1},y_0'\}$ and both $\{v_t\}$ and $\{v_t'\}$ $\stackrel{\text{i.i.d. }}{\sim}N(0,\sigma^2)$.
    
    (iii) The KL-divergence between the distributions of the initial values $y_0,y_0'$ is finite.

Then, given the estimator $\hat\alpha$ produced by the control policy $\pi_t(\alpha^*,\cdot)$, the generalization error on the second system with control policy $\pi_t(\hat\alpha,\cdot)$ has the following convergence rate:
\begin{align*}
    &\frac{1}{T}\sum_{t=0}^{T-1}\mathbb{E}_{y'_{0:T-1}}\bigg(f_t(\beta_0(t),y_t',\pi_t(\hat\alpha,y_t'))-f_t(\beta_0(t),y_t',\pi_t(\alpha^*,y_t'))\bigg)^2=O\bigg(\frac{\log^{c} T}{T}\bigg)\quad a.s.,
\end{align*}
where $c\geq1$ is a constant depending on $n,b$. 
\end{corollary}

\subsection{Theorems on the prediction ability}Next, by combining the above Generalization Lemma \ref{gle} with the meta-LMS prediction Algorithm \ref{pred1} for online adaptation of fine-tuning the drifting parameters in the target system, we can obtain the following theorems.

Denote:
\begin{align*}
    \sigma_{T}^2&\triangleq\frac{1}{T}\sum_{t=0}^{T-1}\mathbb{E}w_{t+1}^2\leq W_{\max}^2,\quad
    L_{0,T}\triangleq\frac{1}{T}\sum_{t=0}^{T-1}L_{0}(t),\\
    \delta_{T}^2&\triangleq\frac{1}{T}\sum_{t=0}^{T-1}\|\beta_{t+1}-\beta_t\|^2\geq\bigg(\frac{1}{T}\sum_{t=0}^{T-1}\|\beta_{t+1}-\beta_t\|\bigg)^2.
\end{align*}

\begin{theorem}\label{thm1}
 Let the assumptions of the Generalization Lemma \ref{gle} as well as Assumption \ref{assa1} holds. If we take the design parameters $d>\frac{A^2}{(1-\gamma)^2}$ and $\lambda<\frac{1}{2(M_f+AB+W_{\max})^2}$ in Algorithm \ref{pred1}, then the average prediction error for the combined two-stage process of offline learning and online adaptation is bounded by:
\begin{equation}\label{thm1bound}
\begin{aligned}
J_T \leq J_{\text{mis}} + J_{\text{opt}} + J_{\text{est}},
\end{aligned}
\end{equation}
where
\begin{align*}
    J_{\text{mis}} &\triangleq \mathcal{N}_d \left( L_1 C \frac{D(P_T \| P_T')}{T^{1 - b_2'}} + L_{0, T} \right), \\
    J_{\text{opt}} &\triangleq \mathcal{N}_d L_1 \varepsilon^*, \\
    J_{\text{est}} &\triangleq \mathcal{N}_d \left( L_1 \frac{C\log(N_1 T)}{N_1 T^{1 - b_2}} + B \delta_T + \delta_T^2 + \frac{B^2}{T} \right) 
                   + \left( \sqrt{C_d} + 1 + \frac{1}{d} \right)^2 \sigma_T^2,
\end{align*}
and where
\(
C_d \triangleq \left( 1 + \frac{A^2}{d} \right)^2 
            \left( 1 + \frac{A^2}{(1 - \gamma)^2 d} \right)
            \left( 1 + \frac{1}{d} \right) \geq 1,
\)
the constant $\mathcal{N}_d$ depends on the design parameter $d$ only and $\mathcal{N}_d = O(d)$; the constant $C$ depends on $n, R_{\mathcal{M}}, b_1, b_1', L, p, \sigma_v$ and the constants $\mathcal{N}_d, C$ can be obtained according to the proofs in Section~\ref{thm1proof} and Appendix~\ref{gleproof}.
\end{theorem}
\begin{remark}
As can be seen from (\ref{thm1bound}), the bound of the prediction error $J_T$ contains three error terms: \textbf{(i)$J_{\text{mis}}$:} the model mismatch containing both the distribution shift characterized by $D(P_T\|P_T')$ and the parameter drift between the source and target systems characterized by $L_1,L_{0,T}$ in Inequality (\ref{transfer}). Note that $L_{0}(t)$ also depends on the generalization error on $\phi_t$, which is a key element linking offline learning in the source system to cross-system drift in the target system; \textbf{(ii)$J_{\text{opt}}$:} the optimization error $\varepsilon^*$ multiplied by $\mathcal{N}_d,L_1$; and \textbf{(iii)$J_{\text{est}}$:} the estimation errors determined by the number of training trajectories $N_1$ and the data-size $T$, by the parameter variation $\delta_T$ in the target system, and by the ``variance'' of the random/deterministic disturbances described by $\sigma_T^2$. Of course, all the three error terms depend on the design parameter $d$ in Algorithm \ref{pred1}.
\end{remark}
\begin{remark}\label{choices}
We remark that various options are available for the selection of $L_1$ and $L_0(t)$ satisfying (\ref{transfer}), and these choices directly influence the prediction error bound. For
guidance on selecting suitable $L_1$ and $L_0(t)$ to further refine the prediction error bound in different scenarios, see Appendix \ref{assbridge} for details.
Furthermore, when multiple source systems with different outer parameters \( \beta_t^{1}, \beta_t^{2}, \cdots, \beta_t^{N_1} \) are considered, the offset term \( L_0(t) \) vanishes under condition that the target parameter \( \beta_t \) lies within the span of the set \( \{ \beta_t^1, \beta_t^2, \cdots, \beta_t^{N_1} \} \), and \( L_1 \) can be taken as the maximum over $t=0$ to $T-1$ of the sum of the squared magnitudes of these \( N_1 \) representation coefficients at each time $t$, as guaranteed by Cauchy-Schwarz inequality similar to Case 2 in Appendix \ref{assbridge}. This holds regardless of whether the offline-learned parameter \( \hat{\alpha} \) converges to the true parameter \( \alpha^* \) and provides theoretical validation that the multi-environment learning strategy, as used in \cite{neuralfly2022neural}, indeed enhances the generalization ability of the offline-learning phase.
\end{remark}
\begin{remark}\label{whyonline}
Theorem \ref{thm1} is established based on (\ref{maxmeta}) in Lemma \ref{predlemma}. Analogous result can be derived from (\ref{minmeta}) with initial weights $w_{0,i}=\frac{1}{N_2}$, which takes the form of a modified version of (\ref{thm1}) wherein the term $\frac{B^2}{T}$ in $J_{\text{est}}$ is replaced by $\frac{\min_{1 \leq i \leq N_2} \mathbb{E}\|\hat{\beta}_{0,i} - \beta_0\|^2}{4T}(\leq \frac{B^2}{T})$, together 
with the inclusion of an additional term $\frac{\log N_2}{\lambda T}$. As shown in FIG. \ref{fig1} of Section \ref{sec5}, employing multiple models significantly enhances the transient response of the system without compromising prediction accuracy. And, for regression vectors of the form \(x_t = (y_t, \cdots,y_{t-p+1},u_t,\cdots, u_{t-q+1})^\tau\)
, analogous conclusions apply \textit{mutatis mutandis} to cases with model-specific regression vectors \(x_{t,i}\), including scenarios where the feedback input \(u_t\) depends on current parameter estimates, as discussed in \cite{lmsguo}, and the proof in Section~\ref{thm1proof} remains valid when substituting \(x_t\) with \(x_{t,i}\), e.g., \cite{yanxia2002stochastic}. We further emphasize that Algorithm \ref{pred1} and its analysis can be generalized to settings with random $\beta_t$, as established in~\cite{lmsguo}. Moreover, as is pointed out in Remark \ref{disandpara}, for many dynamical systems, the KL divergence term in $J_{\text{mis}}$, i.e., \(\frac{D(P_{T} \| P_T')}{T^{1-b_2'}}\),  which quantifies the distribution shift, can be further characterized by parameter drift terms including the source-to-target parameter drift terms $L_1,L_{0,T}$, and the time-varying parameter-induced drift term $\delta_T$, and thus can become particularly small in certain cases. 
\end{remark}
\begin{remark}
Under conditions of no parameter drift or disturbance, i.e., $\delta_T=0,\varepsilon_t=0$ and $w_{t+1}=0$, our projected LMS algorithm guarantees bounded regret, as shown in (\ref{regretlms}) in the proof of Theorem \ref{thm1}. In contrast with the related result in Theorem 1 of \cite{convexmeta}, under the same conditions, \cite{convexmeta} yields a dynamic regret of order $O(T)$ for any fixed hyperparameter. The fundamental reason for this difference lies in the distinct Lyapunov functions used in the analysis. Moreover, following the ideas in \cite{convexmeta}, we can mitigate the trade-off arising from the coefficient $N_d$ and the noise term coefficient as functions of $d$. By incorporating multiple values of design parameters $d_i$ in the meta-algorithm, a tighter upper bound on the prediction error $J_T$ can be derived by using the Lyapunov function in the proof of our Theorem \ref{thm1} in Section \ref{thm1proof}.
\end{remark}

We now show that as $T\to\infty$, the average prediction error will be closed to the optimal value under some typical situations including the case where the disturbance is a martingale difference sequence. 
\begin{theorem}\label{thm1.1}
Under the assumptions of Theorem \ref{thm1}, suppose further that the noise ${w_{t+1}}$ is a bounded martingale difference sequence with respect to ${\mathcal{F}_t'}$ and that $L_1$ is known. If $\delta_{T}^2, L_{0,T} \to 0$, and if either $L_1=0$ or $L_1>0$ with $\frac{D(P_{T}\|P_{T}')}{T^{1-b_2'}}\to0$ as $T\to\infty$, then for any $\varepsilon>0$, by choosing $\varepsilon^*<\frac{\varepsilon}{320L_1}$ (or $\varepsilon^*<\infty$ when $L_1=0$) and $d>\max\{{\frac{A^2}{(1-\gamma)^2},\frac{8A^2W_{\max}^2}{\varepsilon}}\}$, the asymptotic prediction error is bounded as follows:
\begin{equation}
\sigma_{\inf}^2 \leq \liminf_{T\to \infty} J_T \leq \limsup_{T\to \infty} J_T \leq \sigma_{\sup}^2 + \varepsilon,
\end{equation}
where $\mathcal{F}'_t\triangleq\sigma\{\mathcal{\hat\alpha},x_0,x_1,...,x_t,w_1,...,w_{t}\} $, $\sigma_{\inf}^2\triangleq\liminf\limits_{T\to\infty}\sigma_{T}^2 $ and $\sigma^{2}_{\sup}\triangleq\limsup\limits_{T\to\infty}\sigma_{T}^2$.    
\end{theorem}

\begin{remark}
Following the same approach as in the proof of Theorem \ref{thm1.1} in Section \ref{thm1.1proof}, if we additionally assume that either $L_1=0$ or $L_1>0$ with $\varepsilon^*=0$ and $\delta_t$ decays at a known rate, then the optimal prediction performance(i.e., $\varepsilon=0$) can be guaranteed by designing a time-varying hyperparameter $d_t=o(\min\{\frac{1}{\delta_t},t\})$ and $d_t\to\infty$ instead of a fixed $d$.
\end{remark}

\subsection{Further discussions}
Based on the above theorems, let us explain why this two-stage learning method outperforms either purely offline or purely online methods. The simple reason is that this two-stage combined method shares the advantages of both offline and online learning phases. To the best of the authors' knowledge, this natural idea has rarely been explored and elaborated mathematically in theory before.

\textbf{(i) Why the online-adaptation phase is necessary?} Relying solely on offline-learning without online-adaptation typically fails to achieve good prediction performance in real-world systems. More precisely, compared with the relatively ideal source system in the offline-learning phase, the real-world target system may contain unknown changed and changing parameters, besides the negative influence due to the possible non-convergence of the $\hat\alpha$ learned in the offline phase. Luckily, as is pointed out in Chapter 6 of \cite{shibianxitong}, we are capable of addressing these uncertainties using the projected LMS. In contrast, if we do not adaptively adjust the $\hat\beta_t$, and instead directly use the parameters $\beta_t^{0}$ in the offline-learning phase, the errors caused by the model mismatch and disturbance $w_t$ in the system will accumulate(see, FIG. \ref{fig1}-Fixed Parameters). Along with the simulation result, we next provide a theoretical example where the average prediction error $J_T$ becomes unbounded without the online-adaptation phase. Suppose that all the vectors are 1-dimensional for convenience, and that $\beta_t^0\equiv\beta_0\equiv\beta>0$, $\beta_{t+1}=\beta_t+\frac{1}{t+1}$, and $0<\phi_0\leq\mathbb{E}\phi_t^2(\alpha^*,x_t)\leq\phi_1<\infty$ for all $t\geq0$. Additionally, we assume $\hat{\alpha}$ is the NLS estimator of $\alpha^*$(i.e., $\varepsilon^*=0$), the training and new regressor sequences are identically distributed, and Assumption \ref{ass3} holds. In this case, without online-adaptation, even if the true value $\alpha^*$ is known and no external disturbances are present, the prediction error $J_T$ can be lower-bounded as:
$\frac{1}{T}\sum_{t=0}^{T-1}\mathbb{E}(\beta_t\phi_t(\alpha^*,x_t)-\beta\phi_t(\alpha^*,x_t))^2\geq\frac{1}{T}\sum_{t=1}^{T-1}(1+\frac{1}{2}+\cdots+\frac{1}{t})^2\phi_0\to\infty$, as $T\to \infty$. In contrast, if online-adaptation is performed in this case, it is straightforward to verify that $\delta_T^2=O(\frac{1}{T})$ and $B=O(\log T)$. And, since $\beta_t=(1+\frac{1+\frac{1}{2}+\cdots+\frac{1}{t}}{\beta})\beta$, by Case 2 in Appendix \ref{assbridge}, we have $L_1=O(\log T), L_0(t)\equiv0$. Therefore, Theorem \ref{thm1} implies that as $T\to\infty$, $J_T$ will be bounded by $O(\sigma_T^2)$ and the ``O'' constant depends on $d$ only.
    
\textbf{(ii) Why the offline-learning phase is important?} The offline-learning phase enables a reduction in model uncertainty and an improvement in computational efficiency by using historical datasets. More precisely:  \textbf{a).} A well-estimated $\hat{\alpha}$ obtained offline encapsulates valuable prior knowledge that, when integrated into the multi-model meta-algorithm, may improve transient performance particularly when $\hat\beta_{0,i}^\tau \phi_0(\hat\alpha,x_0)\approx \beta_{0}^\tau \phi_0(\alpha^*,x_0)$. This stands in contrast to most purely online adaptation methods that typically suffer from large transient errors due to arbitrary initialization(e.g., \cite{yanxia2002stochastic}), as confirmed in FIG. \ref{fig1}-Single Model.
\textbf{b).} When the dimension of the unknown parameter $\alpha^*$ is quite large, the importance of its offline-learning becomes more apparent, since it can help to reduce the computational burden and increase the flexibility of adaptation in the online-adaptation phase. \textbf{c).} Once we can obtain historical datasets generated from multiple system trajectories/environments, offline learning overcomes the known limitations of single-trajectory system identification (e.g., \cite{duotiaoguijihaochu}) by leveraging multi-trajectory datasets, and at the same time, reduce the offset term $L_{0,T}$ in the bridging inequality, as discussed in Remark 3.11.

\section{Proofs of the main theorems}\label{sec4}

In this section, we present proofs of our main theorems(Theorems \ref{thm1} and \ref{thm1.1}) which are based on \cite{lmsguo,shibianxitong}. We first prove Lemma \ref{predlemma} based on the ideas of \cite{multimodel2024online,predictionlearning,convexmeta}.

\textbf{Proof of Lemma \ref{predlemma}.} By boundedness of $\beta_t$, $\phi(\hat{\alpha}, x_t)$, $\beta_t^\tau \phi(\alpha^*, x_t)$ and $w_{t+1}$ and Lemma 2 in \cite{multimodel2024online}, it follows that $
(y_{t+1} - {y}_{t+1}^\text{pred})^2 \leq (M_f + AB + W_{\max})^2$ is bounded and is $\lambda$-exp-concave for any $0<\lambda<\frac{1}{2(M_f+AB+W_{\max})^2}$. Therefore, by the definition ${y}_{t+1}^{\text{pred}} = \sum_{i=1}^{N_2} w_{t,i} {y}_{t+1,i}^{\text{pred}}$ in Algorithm \ref{pred1} and Jensen inequality,
\begin{align*}
    \exp(-\lambda S_T)&\geq \exp(-\lambda S_{T-1})\sum_{i=1}^{N_2}w_{T-1,i}\exp(-\lambda s_{T-1,i})\\
    &\geq \exp(-\lambda S_{T-2})\sum_{i=1}^{N_2}w_{T-2,i}\exp[-\lambda(s_{T-1,i}+s_{T-2,i})]\\
    &\geq\cdots\geq\sum_{i=1}^{N_2}w_{0,i}\exp\bigg(-\lambda\sum_{t=0}^{T-1}s_{t,i}\bigg),
\end{align*}
where $S_t\triangleq\sum_{j=0}^{t-1}(y_{j+1}-y_{j+1}^\text{pred})^2$, $s_{t,i}\triangleq(y_{t+1}-y_{t+1,i}^{\text{pred}})^2$ and $S_{T,i}\triangleq\sum_{t=0}^{T-1}s_{t,i}$, and the above inequalities hold due to the definition of the weights for $j=1,2,\cdots,T-1$: \[w_{T-j,i}=\frac{w_{T-j-1,i}\exp(-\lambda s_{T-j-1})}{\sum_{i=1}^{N_2}w_{T-j-1,i}\exp(-\lambda s_{T-j-1})}.\]

Furthermore, since
\(\sum_{i=1}^{N_2}w_{0,i}\exp(-\lambda S_{T,i})\geq \min\limits_{1\leq i\leq N_2}\{\exp(-\lambda S_{T,i})\},\)
it is not difficult to see that:
\begin{align*}
     \frac{1}{T}\sum_{t=0}^{T-1}(y_{t+1}-y_{t+1}^{\text{pred}})^2\leq\max\limits_{1\leq i\leq N_2}\bigg\{\frac{1}{T}\sum_{t=0}^{T-1}(y_{t+1}- y_{t+1,i}^{\text{pred}})^2\bigg\}.
\end{align*}

Similarly, since \(\sum_{i=1}^{N_2}w_{0,i}\exp(-\lambda S_{T,i})\geq\max\limits_{1\leq i \leq N_2}\{w_{0,i}\exp(-\lambda S_{T,i})\},\)
we have:
\begin{align*}
   \frac{1}{T}\sum_{t=0}^{T-1}(y_{t+1}-y_{t+1}^{\text{pred}})^2\leq \min\limits_{1\leq i\leq N_2}\bigg\{\frac{1}{T}\sum_{t=0}^{T-1}(y_{t+1}-y_{t+1,i}^{\text{pred}})^2+\frac{1}{\lambda T}\log\frac{1}{w_{0,i}}\bigg\}.
\end{align*}
Taking expectations yields (\ref{minmeta}) immediately from the above inequality.

\subsection{Proof of Theorem \ref{thm1}}\label{thm1proof}

For $i=1,2,\cdots,N_2$ and $k\geq0$, define the Lyapunov function as: \[V_{k,i} = \|\beta_k - \hat{\beta}_{k,i}\|^2 \triangleq \|\tilde{\beta}_{k,i}\|^2\leq 4B^2,\] and denote \(\varphi_k \triangleq \phi_k(\hat\alpha, x_k)\), $\delta_k\triangleq \|\beta_{k+1}-\beta_k\|$.

Since $\|\varphi_k\|\leq A$ and $m_k=\gamma m_{k-1}+\|\varphi_k\|$, we have $\|\varphi_k\|^2\leq m_k^2\leq \frac{A^2}{(1-\gamma)^2}$. 

By boundedness of $\beta_k,\hat\beta_{k+1,i}$,
\begin{align*}
\quad V_{k+1,i} &= \|\beta_{k+1} - \hat{\beta}_{k+1,i}\|^2  \\
&= \|\beta_k + (\beta_{k+1}-\beta_k) - \hat{\beta}_{k+1,i}\|^2\\ 
&\leq \|\beta_k - \hat{\beta}_{k+1,i}\|^2 + 4B\delta_{k} + \delta_k^2, \\
\end{align*}
furthermore, since $\beta_k\in D$, by properties of projection,
\begin{align*}
&\quad\|\beta_k - \hat{\beta}_{k+1,i}\|^2\\&
\leq \left|\left|\left(I - \frac{\varphi_k \varphi_k^\tau}{d+m^2_{k}}\right) \tilde{\beta}_{k,i} - \frac{\varphi_kw_{k+1} + \varphi_k \varepsilon_k}{d+m_k^2}\right|\right|^2  \\
&\leq \|\tilde{\beta}_{k,i}\|^2 -\left(2-\frac{A^2}{d+A^2}\right)\frac{(\tilde{\beta}_{k,i}^\tau \varphi_k)^2}{d+m^2_{k}} + 2 \frac{|\tilde\beta_{k,i}^\tau \varphi_k| \cdot |w_{k+1} + \varepsilon_k|}{d+m^2_{k}} + \frac{A^2(w_{k+1} + \varepsilon_k)^2}{d^2} \\
& \leq \|\tilde{\beta}_{k,i}\|^2 -\left(1-\frac{A^2}{d+A^2}\right)\frac{(\tilde{\beta}_{k,i}^\tau \varphi_k)^2}{d+m^2_{k}}+ \left(\frac{A^2}{d}+1\right) \frac{(w_{k+1} + \varepsilon_k)^2}{d} \\
&= V_{k,i}-\left(1-\frac{A^2}{d+A^2}\right)\frac{(\tilde{\beta}_{k,i}^\tau \varphi_k)^2}{d+m^2_{k}}+ \left(1+\frac{A^2}{d}\right) \frac{(w_{k+1} + \varepsilon_k)^2}{d},
\end{align*}
where the last inequality holds due to $2|x\|y|\leq x^2+y^2$.

Therefore,
\begin{align*}
    \left(1-\frac{A^2}{d+A^2}\right)\frac{(\tilde{\beta}_{k,i}^\tau \varphi_k)^2}{d+m^2_{k}}\leq V_{k,i}-V_{k+1,i}+\left(1+\frac{A^2}{d}\right)\frac{(w_{k+1}+\varepsilon_k)^2}{d}+4B\delta_k+\delta_k^2.
\end{align*}

Since $V_{0,i}=\|\beta_0-\hat\beta_{0,i}\|^2\leq4 B^2$, by summing up from 0 to $T-1$, we can conclude that
\[V_{T,i}+\frac{d}{d+A^2}\sum_{t=0}^{T-1}\frac{(\tilde{\beta}_{t,i}^\tau \varphi_t)^2}{d+m^2_{t}}\leq \frac{A^2+d}{d^2}\sum_{t=0}^{T-1}(w_{t+1}^2+2\varepsilon_tw_{t+1}+\varepsilon_t^2)+4B\sum_{t=0}^{T-1}\delta_t+\sum_{t=0}^{T-1}\delta_t^2+4B^2.\]

Take $d>\frac{A^2}{(1-\gamma)^2}$ to be sufficiently large. It follows that
\begin{equation}\label{regretlms}
\begin{aligned}
\sum_{t=0}^{T-1} (\tilde{\beta}_{t,i}^\tau \varphi_t)^2 
\leq \left(1 + \frac{A^2}{d}\right)^2 \left(1 + \frac{A^2}{(1 - \gamma)^2 d} \right) 
\sum_{t=0}^{T-1}& \left( \left(1 + \frac{1}{d} \right) w_{t+1}^2 + (1 + d) \varepsilon_t^2 \right) \\
&\quad + 16dBT\delta_T + 4dT\delta_T^2 + 16dB^2.
\end{aligned}
\end{equation}

Note that the above inequality holds for all $i=1,2,\cdots,N_2$. By (\ref{maxmeta}) in Lemma \ref{predlemma}, the average prediction error can be upper bounded as follows:
\begin{align*}
    &\quad\frac{1}{T}\sum_{t=0}^{T-1}(y_{t+1}-y_{t+1}^{\text{pred}})^2\\&\leq\max\limits_{1\leq i\leq N_2}\bigg\{\frac{1}{T}\sum_{t=0}^{T-1}\bigg(\beta_{t}^\tau\phi_t(\hat\alpha,x_t)+\varepsilon_{t}+w_{t+1}-\hat\beta_{t,i}^\tau\phi_t(\hat\alpha,x_t)\bigg)^2\bigg\}\\
    &\leq\frac{\sqrt{C_d}d+1+d}{T}\max\limits_{1\leq i\leq N_2}\bigg\{\sum_{t=0}^{T-1}\bigg(\frac{1}{\sqrt{C_d}d}(\tilde{\beta}_{t,i}^\tau\varphi_{t})^2+\varepsilon_{t}^2+\frac{1}{d}w_{t+1}^2\bigg)\bigg\},
\end{align*}
where $C_d\triangleq\left(1+\frac{A^2}{d}\right)^2\left(1+\frac{A^2}{(1-\gamma)^2d}\right)\left(1 + \frac{1}{d} \right)\geq 1$.

Substituting (4.1) into the above inequality and taking the expectation with respect to all randomness yields:
\begin{align*}
   J_T&
    \leq C_d'(B\delta_T+\delta_T^2+\frac{B^2}{T})+C_d''\frac{1}{T}\sum_{t=0}^{T-1}\mathbb{E}\varepsilon_t^2+\left( 
    \sqrt{C_d} + 1 + \frac{1}{d} 
\right)^2\sigma_T^2\\&\leq C_d'(B\delta_T+\delta_T^2+\frac{B^2}{T})+\left( 
    \sqrt{C_d} + 1 + \frac{1}{d} 
\right)^2\sigma_T^2\\&\quad+C_d''L_1\frac{1}{T}\sum_{t=0}^{T-1}\mathbb{E}(\beta_0^\tau(t)(\phi_t(\alpha^*,x_t)-\phi_t(\hat\alpha,x_t))^2+C_d''L_0(t).
\end{align*}

Finally, by Inequality (\ref{transfer}) and Generalization Lemma \ref{gle}, we can get the desired result by direct calculation.
\subsection{Proof of Theorem \ref{thm1.1}}\label{thm1.1proof}

Define the Lyapunov function as $V_{k,i}=\mathbb{E}\|\beta_k-\hat\beta_{k,i}\|^2$. Since $\varepsilon_k,m_k^2,\varphi_k\in\mathcal{F}'_k$, similar to the proof of Theorem \ref{thm1},
we have
\begin{align*}
    &\quad V_{k+1,i}\\&\leq\mathbb{E}\|\beta_k-\hat\beta_{k+1,i}\|^2+4B\delta_k+\delta_k^2\\
    &\leq\mathbb{E}\bigg\|\bigg(I-\frac{\varphi_k\varphi_k^\tau}{d+m_k^2}\bigg)\tilde{\beta}_{k,i}-\frac{\varphi_k(w_{k+1}+\varepsilon_k)}{d+m_k^2}\bigg\|^2+4B\delta_k+\delta_k^2\\
    &\leq\mathbb{E}\|\tilde{\beta}_{k,i}\|^2-(2-\frac{A^2}{d+A^2})\mathbb{E}\frac{(\tilde{\beta}_{k,i}^\tau\varphi_k)^2}{d+m_k^2}+2\mathbb{E}\frac{|\tilde{\beta}_{k,i}^\tau\varphi_k|\cdot|\varepsilon_k|}{d+m_k^2}+\mathbb{E}\frac{A^2(w_{k+1}^2+\varepsilon_k^2)}{d^2}+4B\delta_k+\delta_k^2\\
    &\leq V_{k,i}-(1-\frac{A^2}{d+A^2})\mathbb{E}\frac{(\tilde{\beta}_{k,i}^\tau\varphi_k)^2}{d+m_k^2}+\frac{d+A^2}{d^2}\mathbb{E}\varepsilon_k^2+4B\delta_k+\delta_k^2+\frac{A^2}{d^2}\mathbb{E}w_{k+1}^2,
\end{align*}
where the third inequality holds due to the martingale difference property of $w_{k+1}$.

Summing up from $k=0$ to $T-1$ yields:
\[        \frac{d}{d+A^2}\mathbb{E}\sum_{t=0}^{T-1}\frac{(\tilde{\beta}_{t,i}^\tau\varphi_t)^2}{d+m_t^2}\leq\frac{d+A^2}{d^2}\mathbb{E}\sum_{t=0}^{T-1}\varepsilon_t^2+4BT\delta_T+T\delta_T^2+\frac{A^2}{d^2}TW_{\max}^2+4B^2.\]

Since $d>\frac{A^2}{(1-\gamma)^2}\geq m_t^2$, we have
\begin{equation}
    \begin{aligned}
        \sum_{t=0}^{T-1}\mathbb{E}(\tilde{\beta}_{t,i}^\tau\varphi_t)^2\leq4\sum_{t=0}^{T-1}\mathbb{E}\varepsilon_t^2+8(d+A^2)BT\delta_T+2(d+A^2&)T\delta_T^2+16dB^2\\&+\frac{4A^2}{d}TW_{\max}^2.
    \end{aligned}
\end{equation}

By (\ref{minmeta}) in Lemma \ref{predlemma}, the average prediction error can be upper bounded as follows:
 \begin{align*}
   J_T&\leq\frac{1}{T}\sum_{t=0}^{T-1}\mathbb{E}(\beta_{t}^\tau\phi_t(\hat\alpha,x_t)+\varepsilon_{t}+w_{t+1}-\hat\beta_{t,i}^\tau\phi_t(\hat\alpha,x_t))^2+\frac{\log \frac{1}{w_{0,i}}}{\lambda T}\\
    &=\frac{1}{T}\sum_{t=0}^{T-1}\mathbb{E}(\tilde{\beta}_{t,i}^\tau\phi_t(\hat\alpha,x_t)+\varepsilon_t)^2+\frac{1}{T}\sum_{t=0}^{T-1}\mathbb{E}w_{t+1}^2+\frac{\log \frac{1}{w_{0,i}}}{\lambda T}\\
    &\leq\frac{2}{T}\sum_{t=0}^{T-1}\mathbb{E}(\tilde{\beta}_{t,i}^\tau\phi_t(\hat\alpha,x_t))^2+\frac{2}{T}\sum_{t=0}^{T-1}\mathbb{E}\varepsilon_t^2+\frac{1}{T}\sum_{t=0}^{T-1}\mathbb{E}w_{t+1}^2+\frac{\log \frac{1}{w_{0,i}}}{\lambda T}.
\end{align*}
Since $\varepsilon^*\leq\frac{\varepsilon}{320L_1}$ and $d>\max\{\frac{A^2}{(1-\gamma)^2},\frac{8A^2W_{\max}^2}{\varepsilon}\}$, by (4.2) and Generalization Lemma \ref{gle}, it follows that Theorem \ref{thm1.1} holds.

\section{Numerical Simulation}\label{sec5}
In this section, simulation experiments are conducted to demonstrate that adopting a meta prediction strategy during the online-adaptation phase is beneficial for reducing the prediction error. Furthermore, we demonstrate that, the proposed two-stage offline-online prediction approach maintains satisfactory performance even when the parameter $\hat\alpha$ identified during the initial offline stage deviates considerably from the true value $\alpha^*$.

Consider the real-world target system:
\[y_{t+1}=a_t\sigma_t(bx_t+c)+d_t+w_{t+1},\]
where $\sigma_t(x)=\frac{1}{t+e^{-x}}$ and $W_t$  $\stackrel{\text{i.i.d.}}{\sim} N(0,1)$. $a_t=-50\sigma_t(t)+\frac{1}{t^2}$, $d_t=15\sigma_t(t)+\frac{2}{\log(t+1)}$, $b=1.5$, $c=-0.5$ and $a_t,b,c,d_t$ are unknown beforehand. 

The regression vector sequence $\{x_t\}$ is defined as follows:
\[x_{t+1}=100\sigma_t(x_t)+w_{t+1}',\quad x_0\sim N(10,1),\]
where $w_{t+1}'$ is independent of $w_{t+1}$ and is standard normal.

For simplicity, let $N_1=1$. Parameter estimates of $(b,c)$ are first obtained via offline learning from a dataset generated by the source system. Then, these estimated parameters are fixed in the target system, where Algorithm \ref{pred1} performs continual adaptation of the remaining unknown drifting parameters in an online manner. The perform index of our algorithm in this simulation is the average prediction error: 
\begin{equation}
    \begin{aligned}
        J_{T}\triangleq\frac{1}{T}\sum_{t=0}^{T-1}(y_{t+1}-y_{t+1}^{\text{pred}})^2.
    \end{aligned}
\end{equation}

\textbf{Offline-learning Phase:}
Consider the source system in the offline-learning phase as:
\[y_{t+1}^{off}=a_t^{off}\sigma_t(bx_t^{off}+c)+d_t^{off}+w_{t+1}^{off},\]
where $W_t^{off}$ is independent and identical distributed of $W_t$, $a_t^{off}\equiv20$, $b_t^{off}\equiv-10$, and the training dataset $\{x_t^{off}\}$ is defined as follows:
\[x_{t+1}^{off}=100\sigma_t(x_t^{off})+w_{t+1}^{off'},\quad x_0^{off}\sim N(0,1),\]
where $w_{t+1}^{off'}$ is independent of other random variables and is standard normal.

We then use grid-search in Matlab to get an estimate of $(b,c)$ through least squares: the search domain $\mathcal{M}$ is $[-3.5,6.5]\times[-5.5,4.5]$, and each dimension is uniformly divided into 50 segments.

\textbf{Online-adaptation Phase:}
Fixing the parameters obtained via offline learning and setting $N_2=500,\lambda=10^{-3},d=10^3$ and $D=\{(a,d)|a^2+d^2\leq10^7\}$, we initialize $w_{0,i}=\frac{1}{500}$, $\hat\beta_{0,1}=[10,-10]^\tau$ and $\hat\beta_{0,i}\sim N(-3,1), \text{for }i\geq2$. Algorithm \ref{pred1} is subsequently employed to fine-tune the time-varying parameters. 
\begin{figure}[!htbp]
    \centering
\includegraphics[width=0.5\linewidth, trim=0 8.5cm 0 8cm, clip]{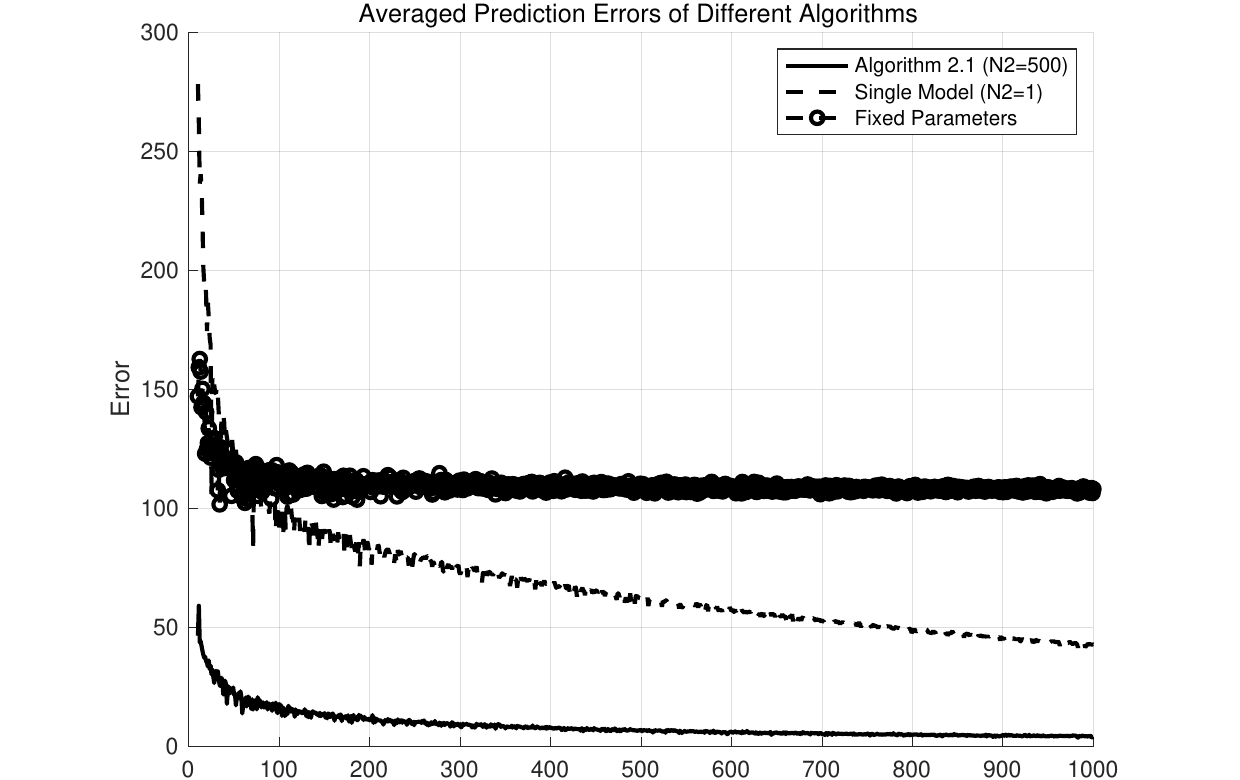}
    \caption{Average prediction error}
    \label{fig1}
\end{figure}
\begin{figure}[!htbp]
    \centering
\includegraphics[width=0.5\linewidth, trim=0 8.5cm 0 8cm, clip]{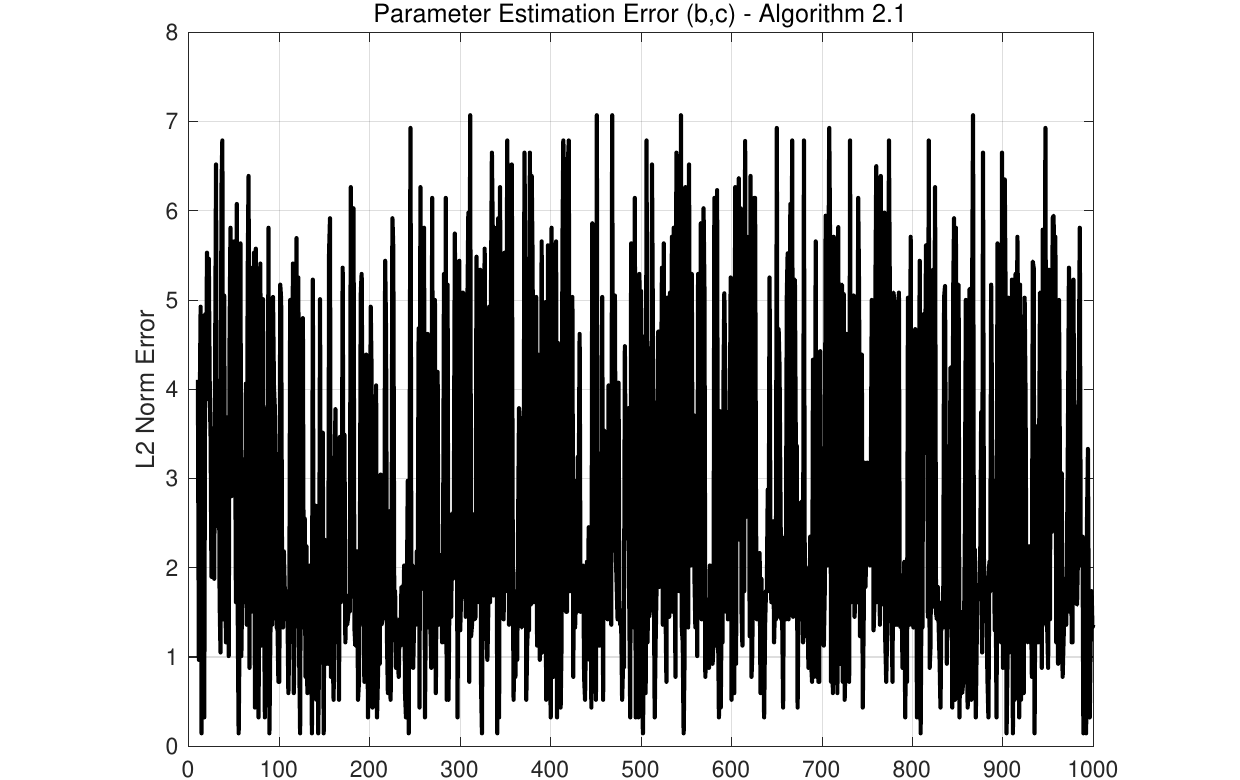}
    \caption{Parameter estimation error of $(b,c)$ in Algorithm 2.1}
    \label{fig2}
\end{figure}

The experimental results demonstrate that the meta approach in Algorithm \ref{pred1} outperforms the single-model method and reduces prediction error(see, FIG. \ref{fig1}), even when the parameter estimate $(\hat b,\hat c)$ fails to converge(see, FIG. \ref{fig2}). Moreover, although a poor initialization leads to a large error in the initial transient response, applying online-adaptation in the second step eventually yields better performance than without adjustment(see, FIG. \ref{fig1}-Single Model and Fixed Parameters).

\section{Concluding Remarks}\label{sec6}
In this paper, we have investigated the prediction performance of a two-stage learning framework combining offline learning and online adaptation for a class of nonlinear stochastic dynamical systems. For the offline phase, we derived a generalization error bound for approximate nonlinear-least-squares estimation under data scenarios involving strong correlation and distribution shift, showing that the generalization error can be quite small under some typical situations. For the online phase, we addressed potential parameter drift in the real-world system by proposing a meta-LMS prediction algorithm that builds upon the offline-trained model. To the best of the authors' knowledge, this paper appears to be the first to demonstrate in theory that this framework can synthesize the complementary strengths of both offline and online learning, resulting in superior prediction performance compared with purely offline or purely online methods. Notably, our results apply under the more general condition of a potentially biased offline estimator. This study confirms that in safety-critical dynamic environments, the integration of offline learning and online adaptation is of paramount importance. It establishes a key theoretical foundation for building reliable learning systems that can withstand various uncertainties, including parameter drift, distribution shift, and external disturbances, etc. Fundamental challenges remain, including developing weaker assumptions on data dependencies and integrating offline reinforcement learning policies into online adaptive control for enhanced performance.

\begin{appendix}
\linespread{0.8}

\section{Remarks on Basic Bridging Inequality (\ref{transfer})}\label{assbridge}
To further refine the upper bound in Theorem \ref{thm1} by selecting appropriate $L_1$ and $L_0(t)$, let us discuss three cases below:

\textbf{Case 1:} $\|\beta_t\beta_t^\tau- K_t\beta_t^0\beta_t^{0\tau}\|\leq B_t$, $K_t$ is scalar, and $\mathbb{E}\|\phi_t(\alpha^*, x_t)-\phi_t(\hat\alpha,x_t)\|^2 \leq M_t,$ for all $t\geq0$. In this case, $\lambda_{\max}(\beta_t\beta_t^\tau-K_t\beta_t^0\beta_t^{0\tau})\leq\| \beta_t\beta_t^\tau-K_t\beta_t^0\beta_t^{0\tau}\|\leq B_t$, thus, by triangle inequality, Inequality (\ref{transfer}) holds provided that $L_1 \geq \max_{0\leq t\leq T-1}|K_t|, L_0(t) \geq B_tM_t$.

\textbf{Case 2:}$\|\beta_t- V_t\beta_t^0\|\leq B_t$, where $V_t\in\mathbb{R}^{m\times m}$ and $\mathbb{E}\|\phi_t(\alpha^*, x_t)-\phi_t(\hat\alpha,x_t)\|^2 \leq M_t,$ for all $t\geq0$. Denote $V'_t(k)\triangleq V_t\beta_t^0\beta_t^{0\tau}V_t^\tau-k\beta_t^0\beta_t^{0\tau}\in\mathbb{R}^{m\times m}$ and its spectral radius $\rho_t(k)\triangleq\rho(V_t'(k))$ for $k\geq0$. Since $\lambda_{max}(\beta_t\beta_t^\tau-V_t\beta_t^0\beta_t^{0\tau}V_t^\tau)\leq\|\beta_t\beta_t^\tau-V_t\beta_t^0\beta_t^{0\tau}V_t^\tau\|\leq B_t(\|\beta_t\|+\|V_t\beta_t^0\|)$, similar to Case 1, by triangle inequality, a feasible choice satisfying Inequality (\ref{transfer}) is $L_1=k,L_0(t)=[\rho_t(k)+B_t(\|\beta_t\|+\|V_t\beta_t^0\|)]M_t.$

Since the offset term $L_0(t)$ can not be eliminated from the finial prediction error(see Theorem \ref{thm1}), we then attempt to determine an $k$ such that $\rho_t(k)\geq0$ is the smallest:

(i) $\beta_t^0$ is an eigenvector of $V_t$ corresponding to an eigenvalue $\lambda_t$. In this case, $\rho_t(|\lambda_t|^2)=0$ is minimal. Thus $L_1=\max_{0\leq t\leq T-1}|\lambda_t|^2$, $L_0(t)=B_tM_t(\|\beta_t\|+\|V_t\beta_t^0\|).$ 

Furthermore, as a special case of (i), if $\beta_{t,i}=\beta_{t,i}^{0}$, $i\neq i_0$, $|\beta_{t,i_0}-\beta_{t,i_0}^{0}|\leq b_t$ and $|\phi_{t}(\alpha^*,x_t)_{i_0}-\phi_{t}(\hat\alpha,x_t)_{i_0}|^2\leq m_t,a.s.$, for all $t\geq0$, then by triangle inequality we can conclude that $|\beta_t^\tau(\phi_t(\alpha^*,x_t)-\phi_t(\hat\alpha,x_t))|\leq |\beta_t^{0\tau}(\phi_t(\alpha^*,x_t)-\phi_t(\hat\alpha,x_t))|+b_tm_t.$ Thus by Cauchy-Schwarz inequality, it follows that $L_1=2, L_0(t)=2b_t^2m_t$ and it suffices to fine-tune parameter $\beta_{t,i_0}$ without adjusting the entire $m$ elements of $\beta_t$ by refining $y_{t+1}'\triangleq y_{t+1}-\sum_{i\neq i_0}\beta_{t,i}^{0}\phi_i(\hat\alpha,x_t)$ in the online-adaptation phase. 

(ii) $V_t\beta_t^0$ and $\beta_t^0$ are linearly independent, thus $\rho_t(k)>0$ for all $k\geq0$. In this case, the spectral radius of $V_t'(k)$ coincides with that of its restriction to $S_t\triangleq span\{\beta_t^0,V_t\beta_t^0\}$. Let $e_1=\frac{\beta_t^0}{\|\beta_t^0\|},e_2=\frac{V_t\beta_t^0-(\beta_t^{0\tau}V_t^\tau e_1)e_1}{\|V_t\beta_t^0-(\beta_t^{0\tau}V_t^\tau e_1)e_1\|}$ form an orthonormal basis of $S_t$. Then the matrix representation of $V_t'(k)$ with respect to $\{e_1, e_2\}$ is
$\begin{pmatrix}
a_t^2 r^2 - k b_t^2 & a_t^2 r_t \sqrt{1 - r_t^2} \\
a_t^2 r_t \sqrt{1 - r_t^2} & a_t^2 (1 - r_t^2)
\end{pmatrix}$, where $a_t\triangleq\|V_t\beta_t^0\|,b_t\triangleq\|\beta_t^0\|$ and $r_t\triangleq \frac{\beta_t^{0\tau}V_t\beta_t^{0}}{ab}<1$. A direct calculation shows that the spectral radius of this $2\times 2$ matrix, when minimized over $k \geq 0$, achieves its minimum at the value $\frac{a_t^2}{2}\left(1-r_t^2+\sqrt{(1-r_t^2)(1+3r_t^2)}\right)$ at $k=\frac{a_t^2r_t^2}{b_t^2}\geq0$.

\textbf{Case 3:} It follows that under additional assumptions, such as the ``excitation'' conditions in Corollary \ref{paraest}, the nonlinear-least-squares estimate $\hat\alpha$ obtained from the offline phase will converge to the true parameter $\alpha^*$ in the mean square sense as $T\to\infty$. Assume that $\beta_t^\tau\phi_t(\alpha,x)$ is uniformly continues  with respect to $\alpha$. By boundedness of $\phi_t(\hat\alpha,x_t),\beta_t$ and $\beta_t^\tau\phi_t(\alpha^*,x_t)$, we can conclude that $\mathbb{E}\Bigg(\beta_t^\tau\bigg(\phi_t(\alpha^*,x_t)-\phi_t(\hat\alpha,x_t)\bigg)\Bigg)^2\to 0$, i.e.,
$L_1=0,L_0(t)\to0$.  

\section{Proof of Generalization Lemma \ref{gle}}\label{gleproof}

Following the approaches in \cite{charis2024rate,singletrajectory2022,activelearning2024}, we first prove Generalization Lemma \ref{gle}. And the 
proof begins with two key lemmas as follows:
\begin{lemma}[see \cite{iidgeneralization,singletrajectory2022}]\label{KLinequality}
 Let \( P \) and \( Q \) be two probability measures defined on a common measurable space \( (\Omega, \mathcal{F}) \), with \( P \ll Q \)(i.e., \( P \) is absolutely continuous with respect to \( Q \)) and let \( F: \Omega \rightarrow \mathbb{R} \) be a measurable function such that \( \mathbb{E}_Q[e^F] < \infty \).

Then, the following inequality holds:
\begin{equation}
\mathbb{E}_P[F]-\log\mathbb{E}_Q[e^F]\leq D(P\|Q).
\end{equation}
\end{lemma}
\begin{lemma}[see \cite{charis2024rate,concentrationinequality,ziemann2024learning}]\label{cdc}For each $t$, let $g_t: \mathcal{M} \times \mathbb{X}^p \to \mathbb{R}$ be such that $0 \leq g_t(\alpha, x) \leq C$, for all $(\alpha, x) \in \mathcal{M}\times \mathbb{X}^p$, for some $C > 0$. Then, for any $\lambda > 0$ and $\alpha \in \mathcal{M}$ we have:
  \begin{align}
    \mathbb{E}\exp\bigg(-\lambda\sum_{t=0}^{T-1}g_t(\alpha,x_t)\bigg)\leq\exp\bigg(-\lambda\sum_{t=0}^{T-1}\mathbb{E}g_t(\alpha,x_t)+\frac{\lambda^2\|\Gamma_{dep}(P_T)\|^2}{2}\sum_{t=0}^{T-1}\mathbb{E}g_t^2(\alpha,x_t)\bigg).
\end{align}  
\end{lemma}
\textbf{Proof of Generalization Lemma \ref{gle}}:
For the new data \(x_{0:T-1}\), denote \[0\leq g_t(\alpha, x_t) \triangleq \|f_t(\alpha, \beta_0(t), x_t) - f_t(\alpha^*, \beta_0(t), x_t)\|^2\leq4L^2R_\mathcal{M}^2.\] 

Take \(F = -\lambda \sum_{t=0}^{T-1} g_t(\alpha, x_t)\) with \(\lambda > 0\) to be determined later. By Lemma \ref{KLinequality}, we have:
\begin{equation}
\begin{aligned}
    &\int - \lambda \sum_{t=0}^{T-1} g_t(\alpha, x_t) dP_{T,i} - \log \int e^{-\lambda \sum_{t=0}^{T-1} g_t(\alpha, x_t)} dP_T'\leq D(P_{T,i} \| P_T').
\end{aligned}    
\end{equation}
By Lemma \ref{cdc}, for the second term on the LHS of (B.3), we have:
\begin{equation}
   \begin{aligned}
&\log\mathbb{E}_{P_T'}\exp\bigg(-\lambda \sum_{t=0}^{T-1} g_t(\alpha, x_t)\bigg) 
\\&\leq \log \left( \exp \left( - \lambda \sum_{t=0}^{T-1} \mathbb{E}_{P_{T}'} g_t(\alpha, x_t)+ \frac{\lambda^2 \| \Gamma_{dep} (P_{T}') \|^2}{2} \sum_{t=0}^{T-1} \mathbb{E}_{P_{T}'} g_t^2(\alpha, x_t) \right) \right)\\
&\leq - \lambda \mathbb{E}_{P_T'} \sum_{t=0}^{T-1} g_t(\alpha, x_t) + \frac{\lambda^2 b_1' T^{b_2'}}{2} \cdot 4L^2R_{\mathcal{M}}^2 \mathbb{E}_{P_T'} \sum_{t=0}^{T-1} g_t(\alpha, x_t).
\end{aligned}         
\end{equation}
Substituting (B.4) into (B.3) and dividing both sides by \(\lambda>0\) yields:
\begin{align*}
    \mathbb{E}_{P_T'}\sum_{t=0}^{T-1}g_t(\alpha,x_t)\leq &\frac{D(P_T \| P_T')}{\lambda}+2\lambda b_1'T^{b_2'}L^2R_{\mathcal{M}}^2\mathbb{E}_{P_T'}\sum_{t=0}^{T-1} g_t(\alpha, x_t)+\mathbb{E}_{P_{T,i}}\sum_{t=0}^{T-1}g_t(\alpha,x_t).
\end{align*}
Without loss of generality, assume \(\mathbb{E}_{P_{T}'}\sum_{t=0}^{T-1} g_t(\hat\alpha, x_t) > 0 \), as the conclusion holds trivially otherwise.\\
If \(D(P_T \| P_T')>0\), by setting \(\lambda = \sqrt{\frac{D(P_T \| P_T')}{2b_1'T^{b_2'} L^2R_{\mathcal{M}}^2 \mathbb{E}_{P_T'}\sum_{t=0}^{T-1}g_t(\hat\alpha,x_t)}}>0\), we then have:
\begin{equation}
    \begin{aligned}
     &\quad\mathbb{E}_{P_T'}\sum_{t=0}^{T-1}g_t(\hat\alpha,x_t) \\&\leq \bigg(\sqrt{2b_1'T^{b_2'} L^2R_{\mathcal{M}}^2 D(P_{T,i} \| P_T')}+ \sqrt{\mathbb{E}_{P_{T,i}} \sum_{t=0}^{T-1} g_t(\hat\alpha, x_t) + 2b_1'T^{b_2'} L^2R_{\mathcal{M}}^2 D(P_{T,i} \| P_T')}\bigg)^2\\
     &\leq 2\mathbb{E}_{P_{T,i}}\sum_{t=0}^{T-1}g_t(\hat\alpha,x_t)+8b_1'T^{b_2'}L^2R_{\mathcal{M}}^2D(P_{T,i}\|P_T');
\end{aligned}
\end{equation}
else, \(D(P_{T,i} \| P_T') = 0\), we have \(\mathbb{E}_{P_{T'}} \sum_{t=0}^{T-1} g_t(\hat\alpha, x_t) = \mathbb{E}_{P_{T,i}} \sum_{t=0}^{T-1}g_t(\hat\alpha, x_t)\), which demonstrates that (B.5) still holds.

Summing over $i$ from $i=1$ to $N_1$ and dividing both sides of (B.5) by \(N_1T\) yields:
\begin{equation}
    \begin{aligned}
     &\frac{1}{T}\mathbb{E}_{P_T'}\sum_{t=0}^{T-1}g_t(\hat\alpha,x_t) \leq \frac{2}{N_1T}\mathbb{E}_{P_{T,i}}\sum_{i=1}^{N_1}\sum_{t=0}^{T-1}g_t(\hat\alpha,x_t)+8b_1'L^2R_{\mathcal{M}}^2\frac{\sum_{i=1}^{N_1}D(P_{T,i}\|P_T')}{N_1T^{1-b_2'}}.
\end{aligned}
\end{equation}

Next, based on the ideas of \cite{charis2024rate}, we analyze the term \(\frac{1}{T}\mathbb{E}_{P_T} \sum_{t=0}^{T-1} g_t(\hat\alpha, x_t)\) in (B.6), where \(x_{0:T-1}\) is independent and identically distributed to the training data \(\hat x_{0:T-1,i}\).

By Lemma \ref{cdc}, we have for any $\lambda>0,\alpha\in\mathcal{M}$:
\begin{align*}
    \mathbb{E}\exp\bigg(-\lambda\sum_{i=1}^{N_1}\sum_{t=0}^{T-1}&g_t(\alpha,\hat x_{t,i})\bigg)
   =\prod_{i=1}^{N_1}\mathbb{E}\exp\bigg(-\lambda\sum_{t=0}^{T-1}g_t(\alpha,\hat x_{t,i})\bigg)\\&\leq\exp\bigg(-\lambda\sum_{i=1}^{N_1}\sum_{t=0}^{T-1}\mathbb{E}_{P_T}g_t(\alpha,\hat x_{t,i})+\frac{\lambda^2b_1T^{b_2}}{2}\sum_{i=1}^{N_1}\sum_{t=0}^{T-1}\mathbb{E}_{P_T}g_t^2(\alpha,\hat x_{t,i})\bigg),
\end{align*}

thus, by Markov inequality, 
\begin{equation}
    \begin{aligned}
      &\quad P\bigg( \frac{1}{T}\sum_{i=1}^{N_1} \sum_{t=0}^{T-1} g_t(\alpha, \hat x_{t,i})\leq \frac{1}{2T} \sum_{i=1}^{N_1}\sum_{t=0}^{T-1} \mathbb{E}_{P_T} g_t(\alpha, \hat x_{t,i})\bigg)\\&\leq \inf_{\lambda>0}P\bigg(\exp\bigg( -\lambda\sum_{i=1}^{N_1} \sum_{t=0}^{T-1} g_t(\alpha, \hat x_{t,i})\bigg)\geq\exp\bigg(-\frac{\lambda}{2} \sum_{i=1}^{N_1}\sum_{t=0}^{T-1} \mathbb{E}_{P_T} g_t(\alpha, \hat x_{t,i})\bigg)\bigg)\\&\leq \inf_{\lambda>0}\frac{\mathbb{E}_{P_T}\exp\bigg( -\lambda\sum_{i=1}^{N_1} \sum_{t=0}^{T-1} g_t(\alpha, \hat x_{t,i})\bigg)}{\exp\bigg(-\frac{\lambda}{2} \sum_{i=1}^{N_1}\sum_{t=0}^{T-1} \mathbb{E}_{P_T} g_t(\alpha, \hat x_{t,i})\bigg)}\leq \exp\left(-\frac{\sum_{i=1}^{N_1}\sum_{t=0}^{T-1} \mathbb{E}_{P_{T,i}} g_t(\alpha, \hat x_{t,i})}{32b_1T^{b_2}L^2R_{\mathcal{M}}^2} \right).   
    \end{aligned}
\end{equation}

Denote \( h_t(\alpha, \hat x_{t,i}) \triangleq f_t(\alpha,\beta_0(t), \hat x_{t,i}) - f_t(\alpha^*, \beta_0(t), \hat x_{t,i}) \) and $h(\alpha,\hat x_{1,i},\hat x_{2,i},\cdots,\hat x_{T-1,i})\triangleq(h_0(\alpha,\hat x_{0,i}),h_1(\alpha,\hat x_{1,i}),\cdots,h_{T-1}(\alpha,\hat x_{T-1,i}))$. Then $\|h_t\|^2= g_t$.

Define the following function spaces:
\begin{align*}
 H^* &= \{ \gamma h(\alpha, X) | \gamma\in [0,1], \alpha \in \mathcal{M}, X=(\hat x_{0,i},\hat x_{1,i},\cdots,\hat x_{T-1,i}) \in \mathbb{X}^{pT} \},\\
H^*_r &= \{ \overline{h} \in H^* | \frac{1}{T} \sum_{t=0}^{T-1} \mathbb{E}_{P_T}\|\overline{h}_t (\alpha, \hat x_{t,i})\|^2 \leq r^2 \},\\
\partial H^*_r &= \{ \overline{h} \in H^* | \frac{1}{T} \sum_{t=0}^{T-1} \mathbb{E}_{P_T}\|\overline{h}_t (\alpha, \hat x_{t,i})\|^2= r^2 \}, \quad r \text{ to be determined later.}    
\end{align*}

For any \( \overline{h} \in H^*\), by (B.7), we have:
\begin{align*}
&\quad P \left( \frac{1}{T} \sum_{i=1}^{N_1}\sum_{t=0}^{T-1} \|\overline{h}_t(\alpha, \hat x_{t,i})\|^2 \leq \frac{1}{2T} \sum_{i=1}^{N_1}\sum_{t=0}^{T-1} \mathbb{E}_{P_T}\|\overline{h}_t (\alpha, \hat x_{t,i})\|^2 \right)
\\&= P \left( \frac{1}{T} \sum_{i=1}^{N_1}\sum_{t=0}^{T-1} \|h_t (\alpha, \hat x_{t,i})\|^2 \leq \frac{1}{2T}\sum_{i=1}^{N_1} \sum_{t=0}^{T-1} \mathbb{E}_{P_T}\|h_t (\alpha, \hat x_{t,i})\|^2 \right)\\
&\leq \exp\left(-\frac{ N_1\sum_{t=0}^{T-1} \mathbb{E}_{P_T} \|h_t(\alpha, \hat x_{t,1})\|^2}{32b_1T^{b_2}L^2R_{\mathcal{M}}^2} \right) \\
&\leq \exp\left(-\frac{ N_1\sum_{t=0}^{T-1} \mathbb{E}_{P_T} \|\overline{h}_t(\alpha, \hat x_{t,1})\|^2}{32b_1T^{b_2}L^2R_{\mathcal{M}}^2} \right).
\end{align*}

Let \( N_r \) be the \( \frac{r}{\sqrt{8}} \) - covering set of \( \partial H^*_r \) in $L^{\infty}$ sense with minimal cardinality. Specifically, a \( \frac{r}{\sqrt{8}} \) - covering set is a set such that for any $\overline{h}\in\partial H_r^*$, there exists $\overline{g}\in N_r\subset \partial H_r^*$, satisfying
$\|\overline{h}-\overline{g}\|_\infty=\max\limits_{0\leq t\leq {T-1}}\|\overline{h}_t-\overline{g}_t\|_\infty\leq\frac{r}{\sqrt{8}}$. Moreover, among all sets possessing this covering property, $N_r$ has minimal cardinality.

By definition of $H^*$, we have for any \(\overline{h} \in N_r, \text{there exists } \overline{\gamma} \in [0,1], \overline{\alpha}\in \mathcal{M}\), s.t. \(\overline{h}=\overline{\gamma}h(\overline{\alpha},X)\).

Denote the event:\[\epsilon = \bigcup_{\overline{\gamma}h(\overline{\alpha},X) \in N_r} \left\{ \frac{1}{N_1T} \sum_{i=1}^{N_1}\sum_{t=0}^{T-1} \overline{\gamma}^2\| h_t (\overline{\alpha}, \hat x_{t,i})\|^2 \leq \frac{1}{2T} \sum_{t=0}^{T-1} \mathbb{E}_{P_T}( \overline{\gamma}^2 \|h_t (\overline{\alpha}, \hat x_{t,1})\|^2) \right\}.\]
We then have:
\begin{align*}
     P(\epsilon) \leq |N_r| \exp\left(-\frac{N_1r^2 T^{1-b_2}}{32 b_1 L^2R_{\mathcal{M}}^2}\right) .
\end{align*}
Similar to (40) of \cite{charis2024rate}, \( |N_r| \) is bounded when $f_t$ is $L$-Lipschitz with respect to $\alpha$ as follows:
\[|N_r|\leq N(\frac{r}{\sqrt{8}},H^*,\|\cdot\|_{\infty})\leq\frac{8\sqrt{8}L R_{\mathcal{M}}}{r} \left( \frac{6\sqrt{8}L R_{\mathcal{M}}}{r} \right)^n\triangleq C(L,n,R_{\mathcal{M}})\frac{1}{r^{n+1}}\] 
Therefore,
\[ P(\epsilon) \leq C( L,n, R_{\mathcal{M}}) \frac{1}{r^{n+1}} \exp \left( -\frac{N_1r^2 T^{1-b_2}}{32b_1L^2R_{\mathcal{M}}^2} \right) .\]
By the definition of \( N_r \), for all \( \overline{h} \in \partial H_r^* \), there exists \( \overline{h'} \in N_r \), s.t.
\(\| \overline{h}(X) - \overline{h'}(X) \|_\infty \leq \frac{r}{\sqrt{8}}, \quad \text{for any } X \in \mathbb{X}^{pT},\)
thus, by the above inequality and $2|x||y|\geq-2x^2-\frac{1}{2}y^2$, we have \( \|\overline{h_t}(\hat x_{t,i})\|^2 \geq \frac{\|\overline{h'}_t^2(\hat x_{t,i})\|^2}{2} - \frac{r^2}{8} \) for any $t=0,1,\cdots,T-1,i=1,2,\cdots,N_1$.
Summing over $i$ and $t$ from $1$ to $N_1$ and $0$ to $T-1$, respectively, and then averaging, yields \[\frac{1}{N_1T} \sum_{i=1}^{N_1}\sum_{t=0}^{T-1}\| \overline{h}_t(\hat x_{t,i})\|^2 \geq \frac{1}{N_1T} \sum_{i=1}^{N_1}\sum_{t=0}^{T-1} \frac{\|\overline{h'}(\hat x_{t,i})\|^2}{2} - \frac{r^2}{8}. \]\\
Furthermore, on \(\epsilon^c\), from the above inequality, we can conclude that:
\[\frac{1}{N_1T}\sum_{i=1}^{N_1} \sum_{t=0}^{T-1} \|\overline{h}_t(\hat x_{t,i})\|^2 \geq \frac{r^2}{4} - \frac{r^2}{8} = \frac{r^2}{8}.\]
Therefore:
\begin{align*}
P\left( \inf_{\overline{h} \in \partial H_r^*} \left( \frac{1}{N_1T} \sum_{i=1}^{N_1}\sum_{t=0}^{T-1} \|\overline{h}_t(\hat x_{t,i})\|^2 - \frac{r^2}{8} \right) \leq 0 \right)
\leq P(\epsilon) \leq C \frac{1}{r^{n+1}} \exp \left( -\frac{N_1r^2 T^{1-b_2}}{32b_1L^2R_{\mathcal{M}}^2} \right).
\end{align*}
Furthermore, for all \(\overline{h} \in H^* \setminus H^*_r\),
\[ \frac{1}{T} \sum_{t=0}^{T-1} \mathbb{E}_{P_T} \|\overline{h}_t(\hat x_{t,1})\|^2 = r'^2 >r^2,\]
this yields \(\frac{r}{r'} < 1\), thus \(\frac{r}{r'}\overline{h}(X) \in \partial H^*_r\). \\
Hence, 
\begin{align*}&\quad P \left( \inf_{\overline{h} \in {H^*\setminus H^*_r}} \left( \frac{1}{N_1T}\sum_{i=1}^{N_1} \sum_{t=0}^{T-1} \|\overline{h}_t(\hat x_{t,i})\|^2 - \frac{1}{8T} \sum_{t=0}^{T-1} \mathbb{E}_{P_T} \|\overline{h}_t(\hat x_{t,1})\|^2 \right) \leq 0 \right)\\&\leq C\frac{1}{r^{n+1}} \exp \left( -\frac{N_1r^2 T^{1-b_2}}{32 b_1 L^2R_{\mathcal{M}}^2} \right).\end{align*}

Denote \(B_r \triangleq \bigg\{\alpha \in \mathcal{M} | \frac{1}{T} \sum_{t=0}^{T-1} \mathbb{E}_{P_T} \|h_t(\alpha, \hat x_{t,1})\|^2 \leq r^2\bigg\}\). For all \(\alpha \in \mathcal{M} \backslash B_r\), we have
\[\frac{1}{T} \sum_{t=0}^{T-1} \mathbb{E}_{P_T} \|h_t(\alpha, \hat x_{t,1})\|^2 > r^2,\]
hence, \[h(\alpha, X) \in H^* \backslash H_r^*.\]
Therefore,
\begin{align*}
    &\quad P \left( \inf_{\alpha \in \mathcal{M} \backslash B_r} \left( \frac{1}{N_1T} \sum_{i=1}^{N_1}\sum_{t=0}^{T-1} \|h_t(\alpha, \hat x_{t,i})\|^2 - \frac{1}{8T} \sum_{t=0}^{T-1} \mathbb{E}_{P_T} \|h_t(\alpha, \hat x_{t,1})\|^2 \right) \leq 0 \right)\\&\leq C\frac{1}{r^{n+1}} \exp \left( -\frac{N_1r^2 T^{1-b_2} }{32 b_1 L^2R_{\mathcal{M}}^2} \right).
\end{align*}

Define the above event as:
\[ \mathcal{A} = \left\{ \inf_{\alpha \in \mathcal{M}\backslash B_r} \left( \frac{1}{N_1T} \sum_{i=1}^{N_1}\sum_{t=0}^{T-1} \|h_t(\alpha, \hat x_{t,i})\|^2 - \frac{1}{8T} \sum_{t=0}^{T-1} \mathbb{E}_{P_T}\|h_t(\alpha, \hat x_{t,1})\|^2\right) \leq 0 \right\} .\]
On \( \mathcal{A}^c \), we have
\[\frac{1}{T} \sum_{t=0}^{T-1} \mathbb{E}_{P_T}\|h_t(\alpha, \hat x_{t,1})\|^2 \leq \frac{8}{N_1T} \sum_{i=1}^{N_1}\sum_{t=0}^{T-1} \|h_t(\alpha, \hat x_{t,i})\|^2 + r^2,\]
on \( \mathcal{A} \),
\[\frac{1}{T} \sum_{t=0}^{T-1} \|h_t(\alpha, \hat x_{t,1})\|^2 \leq 4L^2R_{\mathcal{M}}^2.\]
Therefore,
\begin{align*}
    \frac{1}{T} \sum_{t=0}^{T-1} \mathbb{E}\bigg\|f_t(\alpha,\beta_0(t), \hat x_{t,1}) - f_t(\alpha^*, &\beta_0(t), \hat x_{t,1})\bigg\|^2
    = \frac{1}{T} \sum_{t=0}^{T-1} \mathbb{E}_{P_T}\|h_t(\alpha, \hat x_{t,1})\|^2 \\
    &\leq 4L^2R_{\mathcal{M}}^2 P(A) + \frac{8}{N_1T} \sum_{i=1}^{N_1}\sum_{t=0}^{T-1} \|h_t(\alpha, \hat x_{t,i})\|^2 + r^2.
\end{align*}
Given that the new data $x_{0:T-1}$ and the training data $\hat x_{0:T-1,1}$ are independent and identically distributed, we have
\begin{align*}
    \frac{1}{T} \sum_{t=0}^{T-1} \mathbb{E}g_t(\alpha, x_t)
    =\frac{1}{T} \sum_{t=0}^{T-1} \mathbb{E} g_t(\alpha, \hat x_{t,1})\leq 4L^2R_{\mathcal{M}}^2 P(A) + \frac{8}{N_1T}\sum_{i=1}^{N_1} \sum_{t=0}^{T-1} g_t({\alpha}, \hat x_{t,i}) + r^2.
\end{align*}
Note that the above holds for all \(\alpha \in \mathcal{M}\), specifically, let \(\alpha = \hat{\alpha}\) which is learned from the training datasets.
It yields\begin{equation}
    \begin{aligned}
        \frac{1}{T} \sum_{t=0}^{T-1} \mathbb{E} g_t(\hat{\alpha}, x_t) \leq 4L^2R_{\mathcal{M}}^2 P(A) + \frac{8}{N_1T} \sum_{i=1}^{N_1}\sum_{t=0}^{T-1} g_t(\hat{\alpha}, \hat x_{t,i}) + r^2.
    \end{aligned}
\end{equation}

Furthermore, by Assumption \ref{ass5}:
\[\frac{1}{N_1T} \sum_{i=1}^{N_1}\sum_{t=0}^{T-1} \|\hat y_{t+1,i} - f_t(\hat\alpha, \beta_0(t), \hat x_{t,i})\|^2\leq\frac{1}{N_1T} \sum_{i=1}^{N_1}\sum_{t=0}^{T-1} \|\hat y_{t+1,i} - f_t(\alpha^*, \beta_0(t), \hat x_{t,i})\|^2+\varepsilon^*.\]
By direct computation, we have
\begin{equation}
    \begin{aligned}
            \frac{1}{N_1T}\sum_{i=1}^{N_1} \sum_{t=0}^{T-1} g_t(\hat{\alpha}, \hat x_{t,i}) &\leq \frac{4}{N_1T} \sum_{i=1}^{N_1}\sum_{t=0}^{T-1} \langle v_{t+1,i}, h_t(\hat{\alpha}, \hat x_{t,i})\rangle - \frac{1}{N_1T} \sum_{i=1}^{N_1}\sum_{t=0}^{T-1} g_t(\hat{\alpha}, \hat x_{t,i})+2\varepsilon^*\\
&\triangleq M_{T,N_1}(\hat{\alpha})+2\varepsilon^* \\
&\leq M_{T,N_1}(\mathcal{M}) +2\varepsilon^*,
    \end{aligned}
\end{equation}
where $M_{T,N_1}(\mathcal{M})\triangleq\sup\limits_{\alpha\in \mathcal{M}}M_{T,N_1}(\alpha)$ is the martingale offset term(see, e.g., \cite{martingaleoffset,ziemann2024learning,charis2024rate,singletrajectory2022}) and is a measurable function of $\{\hat x_{0:T-1,i},\hat y_{1:T,i}\}_{i=1}^{N_1}.$

Let $N_\gamma$ be the $\gamma$-covering set of $\{h(\alpha,X)|\alpha\in \mathcal{M},X\in\mathbb{X}^{pT}\}\subset H^*$ with minimal cardinality. Since $h_t$ is bounded, by Assumptions \ref{ass2},\ref{ass4}, similar to (23) of \cite{activelearning2024}, we have for any $\gamma>0$:
\begin{align*}&\quad\mathbb{ E} \left[ M_{T,N_1}(\mathcal{M}) \right] \\&\leq4\gamma\sigma_vp+\gamma^2+\frac{1}{N_1T}\mathbb{ E}\bigg(\sup\limits_{h\in N_\gamma} \frac{4}{N_1T} \sum_{i=1}^{N_1}\sum_{t=0}^{T-1} \langle v_{t+1,i}, h_t(\hat x_{t,i})\rangle - \frac{1}{2N_1T} \sum_{i=1}^{N_1}\sum_{t=0}^{T-1} g_t(\hat x_{t,i})\bigg).\end{align*}\\
For the last term of the expression above, similar to Lemmas 2,3 in \cite{singletrajectory2022}, we have for any $\lambda\in[0,\frac{1}{4\sigma_v^2}]$,
\begin{align*}
    &\quad \mathbb{E}\sup\limits_{h\in N_\gamma}\bigg( \frac{4\lambda}{N_1T} \sum_{i=1}^{N_1}\sum_{t=0}^{T-1} \langle v_{t+1,i}, h_t(\hat x_{t,i})\rangle - \frac{\lambda}{2N_1T} \sum_{i=1}^{N_1}\sum_{t=0}^{T-1} g_t(\hat x_{t,i})\bigg)\\&\leq\log\mathbb{ E}\exp\bigg\{{\sup\limits_{h\in N_\gamma} \frac{4\lambda}{N_1T} \sum_{i=1}^{N_1}\sum_{t=0}^{T-1} \langle v_{t+1,i}, h_t(\hat x_{t,i})\rangle - \frac{\lambda}{2N_1T} \sum_{i=1}^{N_1}\sum_{t=0}^{T-1} g_t(\hat x_{t,i})}\bigg\}\\
    &\leq \log|N_\gamma|\leq (n+1)\log\bigg(C(L,n,R_{\mathcal{M}})\frac{1}{\gamma}\bigg).
\end{align*}
Taking $\lambda=\frac{1}{4\sigma_v^2}$ and combining the above two inequalities yields: 
\begin{equation}
    \begin{aligned}
        \mathbb{ E}[M_{T,N_1}(\mathcal{M})]\leq 4\gamma\sigma_vp+\gamma^2+\frac{4\sigma_v^2(n+1)}{N_1T}\log(\frac{C}{\gamma}).
    \end{aligned}
\end{equation}

Finally, let \(\gamma=\frac{1}{N_1T}, r^2 = \frac{\log(N_1 T)^{\gamma_0}}{N_1T^{1-b_2}} \), where \( \gamma_0 > 16b_1L^2R_{\mathcal{M}}^2(n+3) \) is fixed. By (B.6) and (B.8)-(B.10), we can get the desired result \eqref{maroffset} by direct computation when $N_1\cdot T$ is sufficiently large. For \eqref{G1}, by (92) in \cite{charis2024rate}, Borel-Cantelli Lemma and (B.9), taking $r^2=18b_1L^2R_{\mathcal{M}}^2(n+3)\log(N_1T)/N_1T^{1-b_2}$ yields \eqref{G1}.

\section{Proofs of Corollaries \ref{paraest}, \ref{coro1}, \ref{coro2} and \ref{coro3}}\label{coroproof}

\subsection{Proof of Corollary \ref{paraest}}
(i) Under Condition (\ref{exci1}): By Lemma 2 in \cite{NLSmeasurable1969}, without loss of generality, assume that the NLS estimate $\hat\alpha$ is a measurable function of the training dataset $\{\hat x_{0:T-1,1},\hat y_{1:T,1}\}$. Let the new data $x_{0:T-1}$ be identically distributed to the training data $\{\hat x_{0:T-1,1}\}$ and independent of the NLS estimate $\hat\alpha$. By Generalization Lemma \ref{gle} and the integral form of Mean Value Theorem,
\begin{align*}
    O\bigg(\frac{\log T}{T^{1-b_2}}\bigg)&=\frac{1}{T}\sum_{t=0}^{T-1}\mathbb{E}\bigg\|f_t(\alpha^*,\beta_0(t),x_t)-f_t(\hat\alpha,\beta_0(t),x_t)\bigg\|^2\\
    &= \mathbb{E}(\alpha^*-\hat\alpha)^\tau\bigg(\frac{1}{T}\sum_{t=0}^{T-1}F_t(\alpha^*,\hat\alpha)F_t^\tau(\alpha^*,\hat\alpha)\bigg)(\alpha^*-\hat\alpha)\\&\geq\mathbb{E}\min\limits_{\alpha'\in \mathcal{M}}\lambda_{\min}\bigg(\frac{1}{T}\sum_{t=0}^{T-1}F_t(\alpha^*,\alpha')F_t^\tau(\alpha^*,\alpha')\bigg)\|\alpha^*-\hat\alpha\|^2\\&=\mathbb{E}\Bigg(\min\limits_{\alpha'\in\mathcal{M}}\lambda_{\min}\bigg(\frac{1}{T}\sum_{t=0}^{T-1}F_t(\alpha^*,\alpha')F_t^\tau(\alpha^*,\alpha')\bigg)\Bigg)\mathbb{E}\|\alpha^*-\hat\alpha\|^2,
\end{align*}
where the last equality holds due to the independence of $x_{0:T-1}$ and $\hat\alpha$. Thus we can conclude that $\hat\alpha$ converges to $\alpha^*$ in the mean square sense.

(ii) Under Condition (\ref{exci2}): By (92) in \cite{charis2024rate} and Borel-Cantelli Lemma, we can conclude that the martingale offset term satisfies:
\(\sup_{\alpha\in\mathcal{M}}M_{T,1}(\alpha)=O(\frac{\log T}{T}),a.s.,\)
thus, by (B.9) in the proof of Generalization Lemma \ref{gle}, it follows that 
\[\frac{1}{T}\sum_{t=0}^{T-1}\bigg\|f_t(\alpha^*,\beta_0(t),\hat x_{t,1})-f_t(\hat\alpha,\beta_0(t),\hat x_{t,1})\bigg\|^2=O\bigg(\frac{\log T}{T}\bigg)\quad a.s.,\]
similar to (i), by the integral form of Mean Value Theorem, under Condition (\ref{exci2}), we can conclude that $\hat\alpha$ converges to $\alpha^*$ almost surely.

\subsection{Proof of Corollary \ref{coro1}}
Denote: \[x_t\triangleq(\phi_t^\tau,u_t)^\tau=\bigg(y_t,y_{t-1},\cdots,y
_{t-p+1},u_{t-1},\cdots,u_{t-q+1},\pi_t(\phi_t)+v_{t}\bigg)^\tau,\] 
\[x_t'\triangleq (\phi_t'^\tau,u_t')^\tau=\bigg(y_t',y'_{t-1},\cdots,y'
_{t-p+1},u'_{t-1},\cdots,u'_{t-q+1},\pi_t(\phi_t')+v_{t}'\bigg)^\tau.\]

We first verify that $\{x_t\}$ and $\{x_t'\}$ satisfy Assumptions \ref{ass3}, \ref{ass2} and \ref{ass4}.

By boundedness of $f_t$, $\pi_t$, $\{w,v\}$ and $\{w',v'\}$, denote $s=\max\{p,q-1\}$, we conclude that $\{x_t\}$ and $\{x_t'\}$ are both bounded by $\sqrt{p+q}(W+M_f)$ for $t\geq s$. Hence $\mathbb{X}$ is a compact set in $\mathbb{R}$, and $f_t$ satisfies Assumption \ref{ass2}.

Besides, since $\phi_{t+1}=(y_{t+1},y_{t},\cdots,y_{t-p+2},u_t,u_{t-1},\cdots,u_{t-q+2})^\tau\triangleq F_t'(x_t,w_{t+1})$, by the definition of $x_t$, it yields:
\[x_{t+1}=\begin{pmatrix}
    f_t(\alpha^*,\beta_0(t),x_t)\\
    f_t(\alpha^*,\beta_0(t-1),x_{t-1})\\
    \vdots\\
    f_t(\alpha^*,\beta_0(t-p+1),x_{t-p+1})\\
    \pi_t(\phi_t)\\
    \vdots\\
    \pi_{t-q+2}(\phi_{t-q+2})\\
    \pi_{t+1}(F_t'(x_t,w_{t+1}))\end{pmatrix}+\begin{pmatrix}
    w_{t+1}\\
    w_t\\
    \vdots\\
    w_{t-p+2}\\
    v_t\\
    \vdots\\
    v_{t-q+2}\\
    v_{t+1}
\end{pmatrix},\]
by boundedness of $f_t$ and $\pi_t$, similar to Remark \ref{geophi}, we have that, if $k>s$, for both systems:
\[P_{n,n+k}(x,A)\geq \delta v(A),\]
where $\delta\triangleq \inf_{w\in[-W,W]^{p+q}}p(w)>0$ and $p(w)$ is the density function of the noise, which is a $(p+q)$-dimensional random vector, and $v(\cdot)$ is the Lebesgue measure restricted to $[-W+M_f,W-M_f]^{p+q}$. Thus Assumption \ref{ass3} holds with $b_2=b_2'=0$.

Furthermore, since $\{w_{t+1}\},\{w_{t+1}'\}$ are independent, zero-mean and bounded random variables, Assumption \ref{ass4} holds.

Finally, since $\{x_t'\}$ is a measurable function of $\{w_{t+1}',v_{t+1}',x_0'\}$ and $\{x_t,y_{t+1}\}$ is a measurable function of $\{W_{t+1},v_{t+1},x_0\}$, thus $\{x_t'\}$ is independent of the training dataset.

Next, we address $D(P_T\|P_T')$. Since
\begin{align*}
   x_{t+1}=G_t(x_t,w_{t+1},v_{t+1}),\quad
   x_{t+1}'=G_t(x_t',w_{t+1}',v_{t+1}'),
\end{align*}
where 
\begin{align*}
G_t(x,w,v)\triangleq\bigg(f_t(\alpha^*,\beta_t,x)+w,x_{1},x_{2},\cdots,x_{p-1},x_{p+1}\cdots,x_{p+q-1},\pi_{t+1}(F_t'(x,w))+v\bigg)^\tau,
\end{align*}
therefore, these two independent systems have identical transition kernels, i.e.,
\begin{align*}
    P_{T}&=P(x_0,x_1,\cdots,x_{T-1})=D_0(x_0)\prod_{i=1}^{T-1}P_i(x_{i}|x_{i-1}),\\ P_{T}'&=P'(x_0,x_1,\cdots,x_{T-1})=D_0'(x_0)\prod_{i=1}^{T-1}P_i(x_{i}|x_{i-1}).
\end{align*}

By the chain rule of KL divergence, we have
\begin{align*}
    D(P_{T}\|P_{T}')&=\sum_{i=0}^{T-1}\mathbb{E}_{x_{0:i-1}}D\bigg(P(x_i|x_{0:i-1})\bigg\|P'(x_i|x_{0:i-1})\bigg)\\
    &=D(D_0\|D_0')+\sum_{i=1}^{T-1}\mathbb{E}_{{x_{i-1}}}D\bigg(P_i(x_i|x_{i-1})\bigg\|P_i(x_i|x_{i-1})\bigg)\\
&=D(D_0\|D_0')<\infty.
\end{align*}

Consequently, by Generalization Lemma \ref{gle}, the total expectation of the generalization error diminishes to zero at a rate $\frac{\log T}{T}$.\hfill \qed

\subsection{Proof of Corollary \ref{coro2}}It is obvious that $x_t$ is not independent of the previous sequence, and the training dataset $\{\hat x_{0:T-1},\hat y_{1:T}\}$ is $\{y_{0:T-1},y_{1:T}\}$ and the new data is $y_{T:2T-1}$.

Similar to the proof of Corollary \ref{coro1}, we can conclude that $p(y|x)=p_W(y-F(\alpha^*,x))$, thus for any $ y\in B$, by boundedness of $F$, $\|y-F(\alpha^*,x)\|\leq W$. Thus, there exists $\delta_0>0$, such that the density function of the transition kernel satisfies
\(q(y|x)\geq \delta_0 I\{y\in B\}.\)

Thus, for $y\in B$,
\(p_{T}(y)=\int_X p_{T-1}(x)q(y|x)dx\geq\delta_0.\)

Therefore, the KL divergence of $x_0$, $x_{T}$ can be bounded above as follows:
\begin{align*}
D(P(x_0)\|P
(x_{T}))=\int_Bp_0(x)\log \frac{p_0(x)}{p_{T}(x)}dx\leq\int_{\{x|p_0(x)>1\}}p_0(x)\log p_0(x)dx+\log\frac{1}{\delta_0}<\infty.
\end{align*}
Note that this upper bound is independent of $T$. By the proof of Corollary \ref{paraest} in Appendix \ref{gleproof}, we know that the martingale offset term satisfies $M_{T,1}(\mathcal{M})=O(\frac{\log T}{T}) $ a.s., similar to the proof of Corollary \ref{coro1}, we can conclude that the generalization error over the new data is of order $O(\frac{\log T}{T})$.

\subsection{Proof of Corollary \ref{coro3}}By Remark \ref{noisebound}, we have $v_{t}^2=O(\log t), v_t'^2=O(\log t)$, thus the radius of $\mathbb{X}$ is of order $\sqrt{\log T}$, $L=O(\log T^\frac{b}{2})$. 

Similar to the proof of Corollary \ref{coro1}, we can conclude that $b_2=b_2'=0$.

Besides, for two Gaussian distributions, it yields $D(N(a,\sigma^2)\|N(b,\sigma^2)=\frac{2(a-b)^2}{\sigma^2}$. Note that $D(P_T\|P_T')$ is random since $P_T'$ depends on the random variable $\hat\alpha$. Therefore, by the chain rule of KL divergence, 
\begin{align*}
    &\quad D(P_{T}\|P_{T}')=\sum_{i=0}^{T-1}\mathbb{E}_{y_{0:i-1}}D\bigg(P(y_i|y_{0:i-1})\bigg\|P'(y_i|y_{0:i-1})\bigg)\\
    &=D(D_0\|D_0')+\sum_{i=1}^{T-1}\mathbb{E}_{y_{i-1}}D\bigg(P_i(y_i|y_{i-1})\bigg\|P_i'(y_i|y_{i-1})\bigg)\\
&\leq D(D_0\|D_0')+\sum_{t=0}^{T-2}\mathbb{E}_{y_{t}}\bigg(\frac{2(f_t(\beta_0(t),y_t,\pi_t(\alpha^*,y_t))-f_t(\beta_0(t),y_t,\pi_t(\hat\alpha,y_t))^2}{\sigma^2}\bigg)\\
&=D(D_0\|D_0')+O(\log^cT)\quad a.s.,
\end{align*}
where the inequality holds due to $b_2=b_2'=\varepsilon^*=D(P_T\|P_T)=0$ and (\ref{G1}) in Generalization Lemma \ref{gle}.

Similar to the proof of Corollary \ref{coro1}, the generalization error on the new data $\{y_t'\}$ will converge to 0 as $T\to \infty$.
\end{appendix}	
\bibliographystyle{siamplain}
\bibliography{prediction}

@INPROCEEDINGS{charis2024rate,
  author={Stamouli, Charis and Ziemann, Ingvar and Pappas, George J.},
  booktitle={2024 IEEE 63rd Conference on Decision and Control (CDC)}, 
  title={Rate-Optimal Non-Asymptotics for the Quadratic Prediction Error Method}, 
  year={2024},
  volume={},
  number={},
  pages={5723-5730}}

@article{ziemann2024learning,
      title={Learning with little mixing}, 
      author={Ingvar Ziemann and Stephen Tu},
      journal={arXiv preprint arXiv:2206.08269},
      year={2024},
    
}

@article{neuralfly2022neural,
author = {Michael O’Connell  and Guanya Shi  and Xichen Shi  and Kamyar Azizzadenesheli  and Anima Anandkumar  and Yisong Yue  and Soon-Jo Chung },
title = {Neural-Fly enables rapid learning for agile flight in strong winds},
journal = {Science Robotics},
volume = {7},
number = {66},
pages = {eabm6597},
year = {2022}
}

@INPROCEEDINGS{multimodel2024online,
  author={Vlachos, Anastasios and Tsiamis, Anastasios and Karapetyan, Aren and Balta, Efe C. and Lygeros, John},
  booktitle={2024 IEEE 63rd Conference on Decision and Control (CDC)}, 
  title={Online Residual Learning from Offline Experts for Pedestrian Tracking}, 
  year={2024},
  pages={3390-3395}
  }

@article{yanxia2002stochastic,
    author = {Yanxia Zhang and Lei Guo} ,
    title ={Stochastic adaptive switching control based on multiple models},
    journal ={J. Systems Science and Complexity} ,
    volume = {15},
    pages={18--34},
    year = {2002}
}

@article{singletrajectory2022,
      title={Single Trajectory Nonparametric Learning of Nonlinear Dynamics}, 
      author={Ingvar Ziemann and Henrik Sandberg and Nikolai Matni},
      journal={arXiv preprint arXiv:2202.08311},
      year={2022}
}

@INPROCEEDINGS{activelearning2024,
  author={Lee, Bruce D. and Ziemann, Ingvar and Pappas, George J. and Matni, Nikolai},
  booktitle={2024 IEEE 63rd Conference on Decision and Control (CDC)}, 
  title={Active Learning for Control-Oriented Identification of Nonlinear Systems}, 
  year={2024},
  pages={3011-3018}}

@article{NLSmeasurable1969,
 author = {Robert I. Jennrich},
 journal = {The Annals of Mathematical Statistics},
 number = {2},
 pages = {633--643},
 publisher = {Institute of Mathematical Statistics},
 title = {Asymptotic Properties of Non-Linear Least Squares Estimators},
 volume = {40},
 year = {1969}
}

@book{phimixing1994,
  author    = {Paul Doukhan},
  title     = {Mixing: Properties and Examples},
  year      = {1994},
  publisher = {Springer-Verlag},
  address   = {New York}
}

@article{geocontrol,
author = {Meyn, Sean and Guo, Lei},
year = {2008},
pages = {93 - 108},
title = {GEOMETRIC ERGODICITY OF A DOUBLY STOCHASTIC TIME SERIES MODEL},
volume = {14},
journal = {Journal of Time Series Analysis}
}

@ARTICLE{self1996,
  author={Lei Guo},
  journal={IEEE Transactions on Automatic Control}, 
  title={Self-convergence of weighted least-squares with applications to stochastic adaptive control}, 
  year={1996},
  volume={41},
  number={1},
  pages={79-89}
 }

@article{guo1995,
title = {Convergence and logarithm laws of self-tuning regulators},
journal = {Automatica},
volume = {31},
number = {3},
pages = {435-450},
year = {1995},
author = {Lei Guo}
}

@inproceedings{offlineRLconcen,
 author = {Cortes, Corinna and Mansour, Yishay and Mohri, Mehryar},
 booktitle = {Advances in Neural Information Processing Systems},
 pages = {},
 publisher = {Curran Associates, Inc.},
 title = {Learning Bounds for Importance Weighting},
 volume = {23},
 year = {2010}
}

@ARTICLE{iidgeneralization,
  author={Wu, Xuetong and Manton, Jonathan H. and Aickelin, Uwe and Zhu, Jingge},
  journal={IEEE Transactions on Information Theory}, 
  title={On the Generalization for Transfer Learning: An Information-Theoretic Analysis}, 
  year={2024},
  volume={70},
  number={10},
  pages={7089-7124}}

@book{shibianxitong,
  author    = {L. Guo},
  title     = {Time-Varying Stochastic Systems: Stability and Adaptive Theory},
  edition   = {2nd},
  publisher = {Science Press},
  address   = {Beijing},
  year      = {2020}
}

@article{lmsguo,
author = {Guo, Lei},
title = {On Adaptive Stabilization of Time-Varying Stochastic Systems},
journal = {SIAM Journal on Control and Optimization},
volume = {28},
number = {6},
pages = {1432-1451},
year = {1990},
}

@article{Nature,
    author ={Kaufmann, Elia and Bauersfeld, Leonard and Loquercio, Antonio and Müller, Matthias and Koltun, Vladlen and Scaramuzza, Davide} ,
    title ={Champion-level drone racing using deep reinforcement learning} ,
    journal = {Nature},
    year = {2023},
    volume = {620},
    number = {7976},
    pages = {982--987}
}

@ARTICLE{MPCyingyong,
  author={Mishra, Prabhat K. and Gasparino, Mateus V. and Chowdhary, Girish},
  journal={IEEE Transactions on Automatic Control}, 
  title={Deep Model Predictive Control With Stability Guarantees}, 
  year={2025},
  volume={70},
  number={8},
  pages={5460-5467}}

@book{
Highdimension, place={Cambridge}, series={Cambridge Series in Statistical and Probabilistic Mathematics}, title={High-Dimensional Statistics: A Non-Asymptotic Viewpoint}, publisher={Cambridge University Press}, author={Wainwright, Martin J.}, year={2019}, collection={Cambridge Series in Statistical and Probabilistic Mathematics}
}

@article{baohe,
title = {Identification and adaptation with binary-valued observations under non-persistent excitation condition},
journal = {Automatica},
volume = {138},
pages = {110158},
year = {2022},
author = {Lantian Zhang and Yanlong Zhao and Lei Guo}
}

@article{mpcchen,
author = {Chen, Hui and Guo, Lei},
year = {2023},
title = {Convergence of adaptive MPC for linear stochastic systems},
volume = {66},
journal = {Science China Information Sciences}
}

@inproceedings{
fullbad,
title={Fine-Tuning can Distort Pretrained Features and Underperform Out-of-Distribution},
author={Ananya Kumar and Aditi Raghunathan and Robbie Matthew Jones and Tengyu Ma and Percy Liang},
booktitle={International Conference on Learning Representations},
year={2022}
}

@article{appdnn,
  title={Approximation by superpositions of a sigmoidal function},
  author={George V. Cybenko},
  journal={Mathematics of Control, Signals and Systems},
  year={1989},
  volume={2},
  pages={303-314}
}

@ARTICLE{similarsystem,
  author={Xin, Lei and Ye, Lintao and Chiu, George and Sundaram, Shreyas},
  journal={IEEE Transactions on Automatic Control}, 
  title={Learning Dynamical Systems by Leveraging Data From Similar Systems}, 
  year={2025},
  volume={70},
  number={7},
  pages={4833-4840},
  }

@INPROCEEDINGS{randomsearchwang,
  author={Wang, Xinqiang and Guo, Lei},
  booktitle={2017 36th Chinese Control Conference (CCC)}, 
  title={A new convergent algorithm for online empirical risk minimization}, 
  year={2017},
  pages={11172-11176}
 }

@article{laiwei1982,
title = {Least Squares Estimates in Stochastic Regression Models with Applications to Identification and Control of Dynamic Systems},
 author = {Tze Leung Lai and Ching Zong Wei},
 journal = {The Annals of Statistics},
 number = {1},
 pages = {154--166},
 volume = {10},
 year = {1982}
}

@book{predictionlearning,
author = {Cesa-Bianchi, Nicolò and Lugosi, Gábor},
year = {2006},
title = {Prediction, Learning, and Games},
publisher = {Cambridge University Press},
journal = {Prediction, Learning, and Games}
}

@inproceedings{convexmeta,
author = {Zhang, Lijun and Lu, Shiyin and Zhou, Zhi-Hua},
title = {Adaptive online learning in dynamic environments},
year = {2018},
publisher = {Curran Associates Inc.},
booktitle = {Proceedings of the 32nd International Conference on Neural Information Processing Systems},
pages = {1330–1340},
numpages = {11},
}

@InProceedings{martingaleoffset,
  title = 	 {Learning with Square Loss: Localization through Offset Rademacher Complexity},
  author = 	 {Liang, Tengyuan and Rakhlin, Alexander and Sridharan, Karthik},
  booktitle = 	 {Proceedings of The 28th Conference on Learning Theory},
  pages = 	 {1260--1285},
  year = 	 {2015},
  volume = 	 {40},
  month = 	 {03--06 Jul},
  publisher =    {PMLR}
}

@article{concentrationinequality,
author = {Paul-Marie Samson},
title = {{Concentration of measure inequalities for Markov chains and $\Phi$-mixing processes}},
volume = {28},
journal = {The Annals of Probability},
number = {1},
pages = {416 -- 461},
year = {2000}
}

@ARTICLE{l1baohe,
  author={Zheng, Xin and Guo, Lei},
  journal={IEEE Transactions on Automatic Control}, 
  title={$L_{1}$-Based Adaptive Identification With Saturated Observations}, 
  year={2025},
  volume={70},
  number={9},
  pages={5836-5847},
 }

@article{duotiaoguijihaochu,
title = {Learning linearized models from nonlinear systems under initialization constraints with finite data},
journal = {Automatica},
volume = {180},
pages = {112478},
year = {2025},
issn = {0005-1098},
author = {Lei Xin and Baike She and Qi Dou and George T.-C. Chiu and Shreyas Sundaram}
}

@ARTICLE{senfunction,
  author={Li, Chanying and Guo, Lei},
  journal={IEEE Transactions on Automatic Control}, 
  title={On Feedback Capability in a Class of Nonlinearly Parameterized Uncertain Systems}, 
  year={2011},
  volume={56},
  number={12},
  pages={2946-2951}}

@inproceedings{
mercer,
title={Extending Mercer's expansion to indefinite and asymmetric kernels},
author={Sungwoo Jeong and Alex Townsend},
booktitle={The Thirteenth International Conference on Learning Representations},
year={2025}
}

@article{ representationsong1,
author = { Song,Jian },
title = {High-dimensional functions and manifolds in their optimal representation in low-dimensional visual space},
journal = {Chinese Science Bulletin},
number = {12},
pages = {977-984},
year = {2001},
issn = {0023-074X}
}

@article{representationsong2,
author = {Song, Jian},
title = {Remarks on Representation of Multivariate Functions or Cloud Data in Visualized Low-Dimensional Spaces},
year = {2025},
journal = {Journal of Systems Science and Mathematical Sciences},
volume = {45},
number = {1},
eid = {1},
pages = {1-4}
}

@article{mercerconvergence,
title = {On the speed of uniform convergence in Mercer's theorem},
journal = {Journal of Mathematical Analysis and Applications},
volume = {518},
number = {2},
pages = {126718},
year = {2023},
author = {Rustem Takhanov}
}

@book{Meynmarkov, place={Cambridge}, edition={2}, series={Cambridge Mathematical Library}, title={Markov Chains and Stochastic Stability}, publisher={Cambridge University Press}, author={Meyn, Sean and Tweedie, Richard L. and Glynn, Peter W.}, year={2009}, collection={Cambridge Mathematical Library}}
\end{document}